%% file: main.tex
\documentclass[10pt,twocolumn,letterpaper]{article}

\usepackage[pagenumbers]{cvpr} 
\usepackage[accsupp]{axessibility}  

\input{preamble}
\input{frame_style}

\definecolor{cvprblue}{rgb}{0.21,0.49,0.74}
\usepackage[pagebackref,breaklinks,colorlinks,allcolors=cvprblue]{hyperref}

\def\paperID{37975} 
\def\confName{CVPR}
\def\confYear{2026}

\newcommand{\modelname}[1]{WorldMM}
\title{\modelname{}: Dynamic Multimodal Memory Agent for Long Video Reasoning}

\author{
    Woongyeong Yeo$^{1*}$ \quad
    Kangsan Kim$^{1*}$ \quad
    Jaehong Yoon$^{2\dagger}$ \quad
    Sung Ju Hwang$^{1,3\dagger}$ \\ [1.0mm]
    $^{1}$KAIST \quad $^{2}$NTU Singapore \quad $^{3}$DeepAuto.ai \\ [0.5mm]
    {\tt\small \{wgcyeo, kangsan.kim, sungju.hwang\}@kaist.ac.kr \quad jaehong.yoon@ntu.edu.sg} \\ [0.5mm]
    \normalsize{
        \url{https://worldmm.github.io}
    }
}

\begin{document}
\maketitle

\renewcommand\thefootnote{*}
\footnotetext{Equal contribution; \textsuperscript{$\dagger$}Equal advising}

\input{sec/0_abstract}    
\input{sec/1_introduction}
\input{sec/2_related_works}
\input{sec/3_method}
\input{sec/4_experimental_setup}

\input{sec/5_results_analysis}
\input{sec/6_conclusion}
\input{sec/X_acknowledgement}

{
    \small
    \bibliographystyle{ieeenat_fullname}
    \bibliography{main}
}

\input{sec/X_suppl}

\end{document}

%% file: preamble.tex
\usepackage{makecell}



%% file: frame_style.tex
\newcommand{\whitebox}{%
  \hfill\textcolor{white}{\rule[0.8mm]{1.6mm}{1.6mm}}\hfill%
}

\newcommand{\filmbox}[1]{%
  \IfBeginWith{#1}{'}{%
    \StrGobbleLeft{#1}{1}[\@temp]
    \StrGobbleRight{\@temp}{1}[\filename]%
  }{%
    \edef\filename{#1}%
  }%
  \setlength{\fboxsep}{0pt}%
  \colorbox{black}{%
    \begin{minipage}{2.5cm}
      \rule{0mm}{3.2mm}%
      \whitebox\whitebox\whitebox\whitebox\whitebox%
      \whitebox\whitebox\whitebox\whitebox\null\\
      \null\hfill
        \includegraphics[width=2.4cm]{\filename}%
      \hfill\null\\[0mm]
      \null
      \whitebox\whitebox\whitebox\whitebox\whitebox%
      \whitebox\whitebox\whitebox\whitebox\null
    \end{minipage}%
  }%
}

\newcommand{\whiteboxsmall}{%
  \hfill\textcolor{white}{\rule[0.4mm]{0.8mm}{0.8mm}}\hfill%
}

\newcommand{\filmboxsmall}[1]{%
  \IfBeginWith{#1}{'}{%
    \StrGobbleLeft{#1}{1}[\@temp]
    \StrGobbleRight{\@temp}{1}[\filename]%
  }{%
    \edef\filename{#1}%
  }%
  \setlength{\fboxsep}{0pt}%
  \colorbox{black}{%
    \begin{minipage}{1.3cm}
      \rule{0mm}{1.6mm}%
      \whiteboxsmall\whiteboxsmall\whiteboxsmall\whiteboxsmall\whiteboxsmall%
      \whiteboxsmall\whiteboxsmall\whiteboxsmall\whiteboxsmall\null\\
      \null\hfill
        \includegraphics[width=1.2cm]{\filename}%
      \hfill\null\\[-0.9mm]
      \null
      \whiteboxsmall\whiteboxsmall\whiteboxsmall\whiteboxsmall\whiteboxsmall%
      \whiteboxsmall\whiteboxsmall\whiteboxsmall\whiteboxsmall\null
    \end{minipage}%
  }%
}

%% file: sec/0_abstract.tex
\begin{abstract}

Recent advances in video large language models have demonstrated strong capabilities in understanding short clips. However, scaling them to hours- or days-long videos remains highly challenging due to limited context capacity and the loss of critical visual details during abstraction. Existing memory-augmented methods mitigate this by leveraging textual summaries of video segments, yet they heavily rely on text and fail to utilize visual evidence when reasoning over complex scenes. Moreover, retrieving from fixed temporal scales further limits their flexibility in capturing events that span variable durations. To address this, we introduce WorldMM, a novel multimodal memory agent that constructs and retrieves from multiple complementary memories, encompassing both textual and visual representations. WorldMM comprises three types of memory: episodic memory indexes factual events across multiple temporal scales, semantic memory continuously updates high-level conceptual knowledge, and visual memory preserves detailed information about scenes. During inference, an adaptive retrieval agent iteratively selects the most relevant memory source and leverages multiple temporal granularities based on the query, continuing until it determines that sufficient information has been gathered. WorldMM significantly outperforms existing baselines across five long video question-answering benchmarks, achieving an average 8.4\% performance gain over previous state-of-the-art methods, showing its effectiveness on long video reasoning.

\end{abstract}

%% file: sec/1_introduction.tex
\section{Introduction}
\label{sec:intro}

With the increasing deployment of video large language models (video LLMs)~\cite{gpt5, gemini2.5, qwen3vl, llavavideo} in real-world applications, such as AI glasses and household robots, these models are now required to process and reason over extremely long videos from several hours to even days~\cite{lvbench, egolife, videomme}. Recent works~\citep{videoagent, hippomm, m3-agent} have introduced memory-based approaches that build external memories from abstracted video representations. Such methods allow the model to focus on essential information by retrieving a small number of relevant memories, thereby reducing the number of input tokens. This is a more efficient and effective strategy compared to processing all frames in the video, requiring high computational cost as illustrated in \cref{fig:concept}(a).

Despite their promise, most existing approaches remain highly dependent on textual representations. Typically, each detected event or clip is converted into captions, summaries, or structured text descriptions for downstream retrieval and reasoning~\citep{hippomm, egolife, ego-r1}. Although \citet{m3-agent} incorporates visual inputs when building memory, its use of multimodal features is limited to entity recognition and is not fully exploited during inference (\cref{fig:concept}(b)). Moreover, existing models~\cite{egolife, m3-agent} typically retrieve a fixed number of clips with predetermined durations, such as retrieving three 30-second clips. These rigid architecture designs in video memory agents face the following two major limitations.

\input{fig/fig_concept}

First, they fail to adaptively leverage visual information from videos in conjunction with textual memory during retrieval and generation. Visual details are essential for many real-world tasks requiring attribute recognition, spatial reasoning, or precise scene understanding, while this knowledge cannot be fully represented in text. Meanwhile, as shown in \cref{fig:concept}(c), a fixed strategy that always includes both captions and frames during response generation yields suboptimal results since excessive visual context can even distract the model. Therefore, an adaptive mechanism for selecting multimodal memories is essential for retrieving the most informative references for a given query, which remains unexplored in previous works.

Second, retrieving a fixed number of clips limits the model’s ability to handle queries that require varying temporal scopes. For instance, a question like \textit{``Where did I leave my glasses?"} may require only a few seconds of video, whereas \textit{``What happened in the second half of the soccer match?"} demands a much longer temporal context. Existing approaches retrieve a predetermined length of segments for simplicity~\citep{egolife, m3-agent, hippomm}, which inherently overlooks the diverse temporal scales of real-world events, limiting flexibility. Instead, the retriever should dynamically gather information at multiple temporal scales, combining hour-level summaries with minute-level details as needed.

To fill this gap, we present \modelname{}, a novel memory-based agent that constructs separate textual and visual memories and employs an adaptive retrieval agent to select the optimal memory modality and temporal granularity for each query. The textual memory comprises two components: episodic memory, which stores multiple events across different time scales, and semantic memory, which captures high-level, long-term knowledge such as relationships and habits, organized within knowledge graphs. The visual memory divides a long video into short-term segments indexed within a retrieval corpus, enabling the model to access visual information when required. The retrieval agent iteratively selects the most relevant memory across modalities and timescales, ensuring that the agent retrieves only the information necessary for each query. The proposed multimodal memory selection design therefore prevents the model from being forced to condition on paired yet unnecessary modality memories when retrieving data for a given query, minimizing potential distraction during reasoning.

\input{fig/fig_method}

In addition, \modelname{} is able to retrieve information at varying levels of granularity over the appropriate time range by leveraging multiple graphs operating at different temporal scales, such as seconds, minutes, and hours. When episodic memory is selected for retrieval, the retrieval agent searches each memory to gather potentially relevant information from all temporal levels. The collected candidates are then jointly examined to determine which pieces of information should be used to answer the query. In the end, we dynamically access both short- and long-term video contexts to assemble only the necessary information for reasoning. Furthermore, the model performs retrieval in multiple turns by iteratively selecting memories and queries, thereby expanding the range of possible combinations and allowing adaptive selection of the information for each query.

We evaluate \modelname{} with five long video question-answering benchmarks from hour- to week-long durations. The proposed approach consistently outperforms strong baselines, including long video LLMs and memory-augmented models. Comprehensive ablation studies further demonstrate the effectiveness of our multi-memory, multi-scale design. Specifically, episodic memory enables reasoning over events at multiple timescales, visual memory improves performance on object- and action-centered queries, while semantic memory enhances reasoning over long-term contexts. When all the memories are adaptively integrated, the model achieves the best overall results. These results highlight that our multimodal memory system represents a promising direction toward robust long video reasoning.

%% file: fig/fig_concept.tex
\begin{figure*}[t!]
    \centering
    \includegraphics[width=\textwidth]{fig/src/fig_concept.pdf}
    \vspace{-0.27in}
    \caption{\textbf{Concept Figure.} (a) A day-long video sampled at 1 fps has frames that exceed the context limits of video LLMs. (b) M3-Agent~\citep{m3-agent} relies on textual representation of video, which can underrepresent visual information. (c) EgoRAG~\citep{egolife} retrieves both captions and the corresponding visual frames, but irrelevant frames may distract model. (d) \modelname{} (Ours) constructs multiple memories, incorporating both textual and visual representations, and uses adaptive memory retrieval to effectively leverage multimodal information.
    }
    \label{fig:concept}
    \vspace{-1em}
\end{figure*}

%% file: fig/fig_method.tex
\begin{figure*}[t!]
    \vspace{-0.5em}
    \centering
    \includegraphics[width=\textwidth]{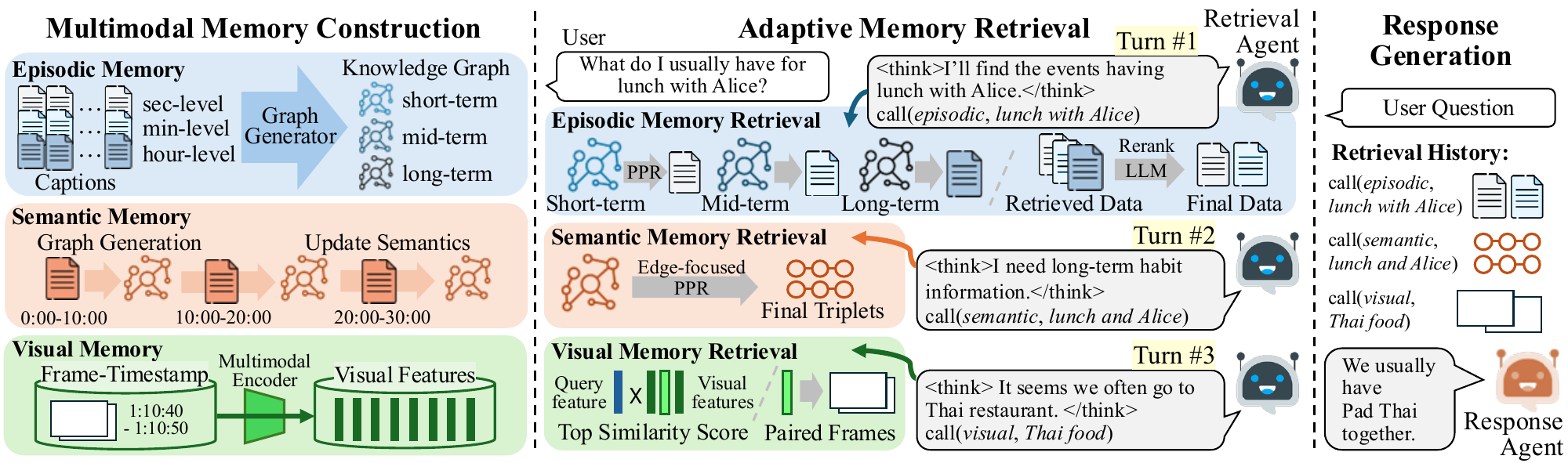}
    \vspace{-0.25in}
    \caption{\textbf{Overview of \modelname{}.} (Left) Multimodal Memory Construction: \modelname{} builds three complementary memories (episodic, semantic, and visual memory) that capture temporal events, long-term relations, and visual details from video streams. (Middle) Adaptive Memory Retrieval: A retrieval agent iteratively selects and integrates relevant information from diverse memories for a given query. (Right) Response Generation: The retrieved content and reasoning history are used by a response agent to produce a grounded response.}
    \label{fig:method}
    \vspace{-1em}
\end{figure*}

%% file: sec/2_related_works.tex
\section{Related Work}
\label{sec:related_work}

\subsection{Long Video Understanding}

Existing video LLMs demonstrate strong understanding capabilities for short videos, and recent research has shifted toward reasoning over longer videos. Current proprietary models, such as GPT-5~\citep{gpt5} and Gemini 2.5~\citep{gemini2.5}, have advanced to minute- or hour-level video understanding~\citep{lvbench, videomme, hippomm, hourvideo, movie} by utilizing extended context lengths. However, these models still incur high computational costs and uniformly sampling every frame is often suboptimal for questions focused on localized events~\citep{toomanyframes,lessismore}.

To address these challenges, several strategies have been explored. Visual token compression~\citep{internlm-xcomposer25, videochat-flash, longllava, everest, inftyvideo, ma-lmm, longvu} improves efficiency by reducing token counts but often loses fine-grained details, limiting the capture of subtle or sparse events. Key frame selection~\citep{videotree, aks} retains only the most informative frames to reduce redundancy but fails to detect relevant frames when video streams are too long and may miss rare events. More recently, reasoning-centric training and inference~\citep{time-r1, video-rts, videomtr, rewatch-r1, conan} have enhanced long-range temporal grounding through reinforcement learning and adaptive test-time scaling, yet still face scalability limits on ultra-long videos over ten hours.

Beyond hour-level videos, emerging benchmarks push video understanding and reasoning toward day- or even week-long continuous recordings~\citep{egolife, ego-r1, mm-lifelong, ma-egoqa}. The aforementioned strategies struggle in these settings due to the massive scale of frames and long-term temporal dependencies. This highlights the necessity for more efficient, context-aware, and scalable approaches to handle extremely long videos.

\subsection{Memory-based Video LLMs}

In order to effectively reason over long videos, retrieval-augmented generation (RAG) based methods that retrieve relevant frames or clips instead of sampled frames have been introduced. They typically retrieve query-relevant information using textual and visual cues, and allow the model to focus on crucial clips~\citep{video-rag, videorag, evrag}. Some recent methods extend this approach by constructing graph structures to encode multimodal interactions across frames~\citep{ren2025videorag, adavideorag, vgent}. However, these models rely on textual representation or a simple similarity score between visual and query features, limiting their ability to perform holistic understanding and complex reasoning over long video sequences.

Beyond naive retrieval-based design, memory-based methods have emerged to construct structured knowledge over video streams. EgoRAG~\citep{egolife} organizes hierarchical textual memories that store events from egocentric video streams in a hierarchical manner, allowing reasoning throughout day-long activities. Ego-R1~\citep{ego-r1} extends this by leveraging vision-centric tools with iterative reasoning to perform long-horizon reasoning. HippoMM~\citep{hippomm} proposes a dual-process memory using semantic summaries with multimodal cues. M3-Agent~\citep{m3-agent} constructs entity-centric long-term memory by processing multimodal contexts and adopts iterative reasoning to retrieve relevant knowledge from the memory. Despite these advances, existing works still struggle to fully integrate multimodal information and to dynamically retrieve knowledge across varying temporal scales to handle complex, long video scenarios.

%% file: sec/3_method.tex
\section{\modelname{}}
\label{sec:method}

We introduce \modelname{}, a novel framework that leverages both textual and visual contexts of video streams to build a multimodal memory for comprehensive understanding and reasoning over long videos. As illustrated in \cref{fig:method}, the model operates in three stages: multimodal memory construction, adaptive memory retrieval, and response generation. Given a long video stream, \modelname{} first builds multiple memories, including two textual memories and one visual memory (\cref{subsec:multimodal_memory_construction}).
Next, a retrieval agent iteratively collects query-relevant information from different memories and timescales until sufficient knowledge is gathered to answer the question (\cref{subsec:adaptive_memory_retrieval}).
Finally, the query is fed into a response agent along with the retrieved contents and retrieval history to generate a response (\cref{subsec:response_generation}).

\subsection{Multimodal Memory Construction}
\label{subsec:multimodal_memory_construction}
As we described in \cref{sec:intro}, an effective memory agent for long-form video understanding must address two key requirements: 1) \textit{adaptively leveraging visual information alongside text memory}, and 2) \textit{retrieving knowledge across diverse temporal ranges.}
To achieve this, \modelname{} constructs three types of memory, each encoding complementary video knowledge across diverse modalities. Episodic memory captures diverse events over multiple dynamic timescales, semantic memory incrementally updates high-level relational knowledge, and visual memory preserves spatial and appearance details. Together, they form a comprehensive multimodal memory that supports episodic retrieval, semantic reasoning, and visually grounded understanding of long-form videos.

\paragraph{Episodic Memory Construction}
Episodic memory consists of multiple textual graphs, each of which encodes events at different temporal resolutions. 
Before constructing the graphs, we first perform fine-grained captioning on the unit temporal scale $t_0$. We divide the video into short segments of length $t_0$, each converted into a caption using a video LLM.
Most existing approaches rely on a fixed temporal scale during memory construction~\citep{egolife, m3-agent}, overlooking the diverse spans of real-world events. In contrast, we introduce a multi-scale memory composed of multiple temporal resolutions that flexibly encodes information with different levels of density:
\begin{equation}
    \mathcal{T}=\{t_0, t_1, \dots, t_N\}, \quad t_0<t_1<\cdots<t_N.
\end{equation}
For each temporal scale $t_i\in \mathcal{T}$, the video is partitioned into non-overlapping segments of length $t_i$. The segments are captioned and transformed into factual triplets (entity-action-entity) to construct a knowledge graph (KG) $G_{t_i}$.
Finally, episodic memory is represented as a set of KGs:
\begin{equation}
    \mathcal{M}_{e} = \{ G_{t_0}, G_{t_1}, \dots, G_{t_N}\}.
\end{equation}
This multi-scale episodic memory enables temporally grounded reasoning that spans both fine-grained event details and long-range narrative understanding.

\paragraph{Semantic Memory Construction}

Semantic memory captures long-term, evolving knowledge about relationships and habits within a video.
Since episodic graphs are constructed from independent events, they fail to preserve continuity across distant scenes and cannot capture high-level knowledge. Semantic memory, on the contrary, maintains an evolving graph that continuously integrates relational and habitual knowledge over time.

To build this continually updating memory, we first split the input video into coarse segments with a fixed timescale $t_s$. Textual captions are generated for each segment and converted into semantic triplets $T_{t_s}^k$, focusing on conceptual knowledge rather than event-specific details.
These triplets are incrementally integrated into an evolving semantic graph through a consolidation process that merges new knowledge while preserving stable relationships. Specifically, embedding-based similarity is first used to identify overlapping or conflicting triplets between the current graph $G_{t_s}^k$ and the newly extracted triplets $T_{t_s}^{k+1}$. The matched triplets are then provided to an LLM along with the new triplets, which determines outdated or conflicting triplets $T_{\text{remove}}$ and triplets that should be revised or added $T_{\text{update}}$. Formally, the consolidation process can be represented as:
\begin{equation}
    \text{Consolidate}(G_{t_s}^k, T_{t_s}^{k+1}) = \big(G_{t_s}^k \setminus T_{\text{remove}}\big) \cup T_{\text{update}}.
\end{equation}
The resulting semantic memory is a continuously evolving KG $\mathcal{M}_s=G_{t_s}^M$, where $M$ denotes the final segment index, capturing the video’s long-term knowledge.

\paragraph{Visual Memory Construction}

Visual memory captures rich visual details that text cannot fully convey, including detailed object appearances, scene dynamics, and precise spatial context.
\modelname{} explicitly constructs a visual memory to ground reasoning in visual evidence. 
We consider two scenarios in which visual memory is invoked, when the retrieval agent searches for scenes associated with a specific keyword, and when the agent has timestamps identified during preceding retrieval steps to inspect the corresponding frames. Therefore, we adopt two complementary strategies for building visual memory: feature-based retrieval via natural language query, and timestamp-based retrieval for precise temporal grounding.

Specifically, we partition each video into short, fixed-length segments of duration $t_v$, encoding each segment $V_{t_v}^k$ into a visual feature $f_v^k$ using a multimodal encoder, forming a feature-based visual memory as a set of embeddings:
\begin{equation}
    \mathcal{M}_{v}^{f} = \{f_v^{1}, f_v^{2}, \dots, f_v^{L} \}.
\end{equation}
In parallel, to support timestamp-based retrieval, each frame is paired with its corresponding timestamp:
\begin{equation}
    \mathcal{M}_v^I = \{(t_i, I_i) \mid I_i = V(t_i),\ t_i \in [0, \text{len}(V)]\}.
\end{equation}
This allows direct access to visual evidence at specific moments in the video.
Finally, the complete visual memory integrates both components $\mathcal{M}_v = \mathcal{M}_v^f \cup \mathcal{M}_v^I$ by combining feature-level embeddings and frame-level indices.

\subsection{Adaptive Memory Retrieval}
\label{subsec:adaptive_memory_retrieval}

In this section, we present how \modelname{} dynamically retrieves the most relevant multimodal memories from the appropriate temporal scope for a given query.

\paragraph{Retrieval Agent}

Reasoning over long-form videos requires integrating heterogeneous information from multiple memory sources. To handle this, the retrieval agent iteratively decides which memory to access and what query to issue, conditioned on the user question and retrieval history. Leveraging the distinct characteristics of each memory component, it adaptively selects the most relevant source and formulates modality-specific queries~\citep{adaptiverag, universalrag}. Through successive iterations, the agent progressively refines its retrieval strategy and constructs better knowledge collection.

Formally, we define the retrieval agent $\mathcal{R}$ as a multimodal reasoning module that iteratively selects a memory source and formulates a corresponding query. At each iteration $i$, $\mathcal{R}$ takes as input the user query $q$ and the set of previous retrieval histories $r_{<i}=\{r_1, \dots, r_{i-1}\}$, and outputs either a memory–query pair or a \texttt{STOP} signal:
\begin{equation}
    \mathcal{R}(q, r_{<i}) =
    \begin{cases}
        (m_i, q_i) & \small\text{if $r_{<i}$ insufficient and $i\le N$},\\
        \texttt{STOP} & \small\text{otherwise},
    \end{cases}
\end{equation}
where $m_i \in \{\mathcal{M}_e, \mathcal{M}_s,\mathcal{M}_v\}$ and $N$ denotes the maximum number of iterations.
If the retriever outputs a memory–query pair $(m_i, q_i)$, it retrieves the relevant information from the memory $m_i$ with search query $q_i$ and proceeds to the next iteration with the updated context $r_{\le i}$.
When the retriever outputs \texttt{STOP}, it indicates that sufficient information has been collected. The iterative process then terminates, and all retrieved results $\{r_1, \ldots, r_n\}$ are passed to the response agent for the final response generation.

\paragraph{Episodic Memory Retrieval}
Episodic memory retrieval is guided by a query $q$ provided by the retriever, which specifies the desired information from episodic memory. The main challenge lies in determining the appropriate temporal scope, as episodic memory contains multiple graphs covering different temporal ranges. \modelname{} adopts a coarse-to-fine, multi-timescale retrieval strategy. Specifically, for each temporal graph $G_{t_i}$, the model first retrieves the top-$k$ candidate captions using a graph-based retrieval framework guided by the Personalized PageRank (PPR) score and the query $q$, following ~\citet{hipporag}. Subsequently, an LLM serves as a cross-scale reranker, jointly analyzing the query and candidates across all timescales. It then selects the most relevant temporal range and refines the retrieved content, producing the final top-$m$ captions as output. By retrieving from multi-scale memory, the model leverages both coarse temporal context and fine-grained details.

\paragraph{Semantic Memory Retrieval}
The semantic memory, also represented as a graph, is queried using a PPR-based retrieval algorithm. In contrast to episodic memory retrieval which operates over nodes and their temporal structures, semantic retrieval focuses on relational knowledge encoded as edges between entities. Since the standard PPR score measures node-level relevance, we adapt it for edge-based reasoning by assigning each edge a score equal to the sum of the PPR values of its two connected nodes. The top-$k$ triplets corresponding to the highest-scoring edges are then selected as the final retrieved results.

\paragraph{Visual Memory Retrieval}
Following \cref{subsec:multimodal_memory_construction}, visual memory retrieval operates in two complementary modes: feature-based search and timestamp-based access. In the feature-based mode, the retrieval agent formulates a query $q$, encodes it into a text feature $f_t$ using a multimodal encoder, and retrieves the top-$k$ relevant video segments from $\mathcal{M}_v^f$ based on the cosine similarity between $f_t$ and the visual features. In the timestamp-based mode, when specific temporal ranges are identified, typically following episodic retrieval, the corresponding frames are directly fetched from $\mathcal{M}_v^I$. By combining these two modes, \modelname{} enables flexible and effective access to visual information at both semantic and temporal levels.

\subsection{Response Generation}
\label{subsec:response_generation}
Finally, once the retrieval agent determines that sufficient information has been gathered, the retrieval process is terminated. The retrieval history, including the selected memories, their corresponding queries, and the retrieved results, is then passed to the response agent along with the original user query. The response agent generates the final answer by grounding its response in the retrieved information. This clear separation between the retriever and the responder allows each component to focus on its respective objective, ensuring effective retrieval and response generation.

%% file: sec/4_experimental_setup.tex
\section{Experiment Results}
\label{sec:experiments}

\subsection{Experimental Setup}

\paragraph{Datasets and Metrics}
We assess the performance of \modelname{} across five benchmarks that require reasoning over long videos. EgoLifeQA~\citep{egolife} and Ego-R1 Bench~\citep{ego-r1} contain week-long videos, and HippoVlog~\citep{hippomm} features vlog-style content, requiring comprehension of audio and visual streams. We also assess general video understanding on LVBench~\citep{lvbench} and Video-MME (long)~\citep{videomme} with hour-level videos. All benchmarks consist of multiple-choice questions, with accuracy used as the evaluation metric. Please see additional dataset details in the \cref{sec:appendix_dataset}.

\paragraph{Baselines}
We compare \modelname{} against a comprehensive set of baselines spanning base video LLMs, long video understanding models, RAG systems, and memory-based models.
Base video LLMs include GPT-5~\cite{gpt5}, Gemini 2.5 Pro~\cite{gemini2.5}, and Qwen3-VL-8B-Instruct~\cite{qwen3vl}, while long video understanding models include VideoChat-Flash~\cite{videochat-flash}, Time-R1~\cite{time-r1}, and Video-RTS~\cite{video-rts}, which all use uniformly sampled frames within their input capacity.
We further evaluate RAG approaches, including text retrieval methods like LightRAG~\cite{lightrag} and HippoRAG~\cite{hipporag}, which retrieve video captions, and Video-RAG~\cite{video-rag}, which retrieves relevant clips.
Finally, we compare with memory-based frameworks for long video reasoning, including EgoRAG~\cite{egolife}, Ego-R1~\cite{ego-r1}, HippoMM~\cite{hippomm}, and M3-Agent~\cite{m3-agent}.

\paragraph{Implementations Details}
We adopt VLM2Vec-V2~\citep{vlm2vec} as a multimodal encoder for visual memory retrieval.
During the memory construction, GPT-5-mini is used for building episodic and semantic memories. We experiment ours with two video LLMs, GPT-5 and Qwen3-VL-8B-Instruct, which serve as the retrieval and response agent, respectively denoted as \modelname{}-GPT and \modelname{}-8B. We apply dataset-specific timescales for temporal segmentation within the episodic memory. For example, we use 30-second, 3-minute, 10-minute, and 1-hour intervals for EgoLifeQA. Configurations for other benchmarks, more experimental details, and the prompts are provided in \cref{sec:appendix_implementation_details}.

\input{tab/tab_main}

%% file: tab/tab_main.tex
\begin{table}[t]
\centering
\setlength{\tabcolsep}{3pt}
\renewcommand{\arraystretch}{0.95}
\caption{Performance of \modelname{} with various baselines across long video QA benchmarks. ``--'' denotes a proprietary backbone.}
\vspace{-0.05in}
\label{tab:main_results}
\resizebox{\linewidth}{!}{
\begin{tabular}{lccccccc}
\toprule
\textbf{Model} & &
\makecell{\textbf{EgoLife} \\ \textbf{QA}} &
\makecell{\textbf{Ego-R1} \\ \textbf{Bench}} &
\makecell{\textbf{Hippo} \\ \textbf{Vlog}} &
\makecell{\textbf{LV} \\ \textbf{Bench}} &
\makecell{\textbf{Video-MME} \\ \textbf{(L)}} &
\textbf{Avg.} \\
\midrule
\multicolumn{5}{l}{\textbf{\textit{Base Models}}} \\
Qwen3-VL-8B~\citep{qwen3vl} & 8B & 38.6 & 35.7 & 74.4 & 48.3 & 61.0 & 51.6 \\
Gemini 2.5 Pro~\citep{gemini2.5} & -- & 46.4 & 46.7 & 72.0 & 57.0 & 55.7 & 55.6 \\
GPT-5~\citep{gpt5} & -- & 48.6 & 46.3 & 75.7 & 60.4 & 74.3 & 61.1 \\
\midrule
\multicolumn{5}{l}{\textbf{\textit{Long Video LLMs}}} \\
VideoChat-Flash~\citep{videochat-flash} & 7B & 34.2 & 42.7 & 58.0 & 33.2 & 44.1 & 42.4 \\
Time-R1~\citep{time-r1} & 7B & 48.8 & 48.0 & 54.6 & 31.1 & 37.6 & 44.0 \\
Video-RTS~\citep{video-rts} & 7B & 48.2 & 47.4 & 59.0 & 39.8 & 47.9 & 48.6 \\
\midrule
\multicolumn{5}{l}{\textbf{\textit{RAG-based Video LLMs}}} \\
LightRAG~\citep{lightrag} & -- & 48.8 & 52.3 & 47.4 & 30.4 & 46.6 & 45.1 \\
HippoRAG~\citep{hipporag} & -- & 59.6 & 56.0 & 63.2 & 54.0 & 52.1 & 57.0 \\
Video-RAG~\citep{video-rag} & -- & 55.4 & 49.7 & 65.1 & 33.1 & 55.4 & 51.7 \\
\midrule
\multicolumn{5}{l}{\textbf{\textit{Memory-based Video LLMs}}} \\
EgoRAG~\citep{egolife} & -- & 52.0 & 49.0 & 57.5 & 32.2 & 41.1 & 46.4 \\
Ego-R1~\citep{ego-r1} & 3B & 53.0 & 52.0 & 58.8 & 34.1 & 42.7 & 48.1 \\
HippoMM~\citep{hippomm} & -- & 54.6 & 53.0 & 71.9 & 38.2 & 41.6 & 51.8 \\
M3-Agent~\citep{m3-agent} & 7B & 53.5 & 52.0 & 65.5 & 49.3 & 55.3 & 55.1 \\
\midrule
\rowcolor{myblue} \multicolumn{8}{l}{\textbf{\textit{\modelname{} (Ours)}}} \\
\rowcolor{myblue} \textbf{WorldMM-8B} & 8B & 56.4 & 52.0 & 69.7 & 55.4 & 66.0 & 59.9\\
\rowcolor{myblue} \textbf{WorldMM-GPT} & -- & \textbf{65.6} & \textbf{65.3} & \textbf{78.3} & \textbf{61.9} & \textbf{76.6} & \textbf{69.5} \\
\bottomrule
\end{tabular}
}
\vspace{-0.1in}
\end{table}

%% file: sec/5_results_analysis.tex
\subsection{Main Results}

\cref{tab:main_results} presents the evaluation results of the proposed \modelname{} and baseline models. \modelname{} significantly outperforms all baselines across various long video understanding benchmarks. In particular, \modelname{}-GPT achieves an average score of 69.5\%, exceeding the strongest baseline by 8.4\%. Compared with base models, both variants of our model surpass their corresponding baselines by more than 8\% on average, highlighting the effectiveness of our framework in leveraging strong reasoning capabilities without requiring full video processing.
On the other hand, models in the long video LLM category show the weakest performance, with all results falling below 50\% on EgoLifeQA and Ego-R1 Bench, indicating that these approaches are not effective in days-long videos.

Meanwhile, retrieval- and memory-based approaches, including ours, achieve scores mostly above 52\% on EgoLifeQA and Ego-R1 Bench, suggesting that selective retrieval of relevant segments is more effective for long video understanding than processing full video sequences.
Compared with other retrieval-based models such as HippoRAG and HippoMM, which also rely on GPT backbones, our model achieves markedly higher accuracy on average (69.5\% vs. 57.0\% and 51.8\%). These demonstrate that integrating textual and visual memory and adaptively selecting temporal scopes are crucial for effective video reasoning.

\input{fig/fig_memtype}
\input{tab/tab_memory_type}
\input{fig/fig_qualitative}
\input{fig/fig_tables}
\input{fig/fig_figures}

\subsection{Efficacy of Multimodal Memory}

To examine the contribution of each memory in our framework, we perform an ablation study that varies the composition of available memories. The evaluation results, summarized in \cref{tab:memory_type}, show a consistent improvement in performance as additional memories are incorporated. This finding confirms that different memories capture complementary forms of knowledge. All following experiments in this section are conducted using \modelname{}-GPT.

\paragraph{Effect of Episodic Memory}
To examine the performance differences arising from the retrieved data modality in \modelname{}, we evaluate models using only episodic memory (E) and only visual memory (V), and report the results in \cref{tab:memory_type}. Using only episodic memory shows 20\% higher performance than using only visual memory on average. This is mostly because textual information can be more readily organized into a graph, which enables effective retrieval, while indexing visual frames into a structured representation remains challenging.

\paragraph{Effect of Visual Memory}
Visual memory plays a particularly important role in categories that demand perceptual understanding, such as object recognition or action interpretation. In \cref{tab:memory_type}, visual memory significantly enhances accuracy in categories like EntityLog and EventRecall of EgoLifeQA and Ego-R1 Bench, as well as Visual and Audio+Visual of HippoVlog. The full configuration (E+S+V) surpasses the non-visual configuration (E+S) by an average margin of 4.2\%. This improvement arises because visual information preserves spatial and perceptual details that are difficult to represent in text. As shown in \cref{fig:qualitative}(a), when relying solely on episodic memory, the model fails to capture object details such as the type of baked item, leading to an incorrect response. In contrast, visual memory provides access to corresponding frames that contain a complete scene, enabling accurate interpretation of objects and activities.

\paragraph{Effect of Semantic Memory}
Semantic memory proves the most beneficial for categories that require reasoning over long-term dependencies and abstract relationships. This effect is evident in the HabitInsight and RelationMap categories of EgoLifeQA and Ego-R1 Bench. The model equipped with full memory achieves 76.9\% accuracy in HabitInsight, representing a 23\% improvement over the setting without semantic memory (E+V). This substantial gain indicates that semantic memory serves as a structured knowledge base that captures relational or habitual knowledge accumulated over time. The qualitative example in \cref{fig:qualitative}(b) illustrates this behavior, where episodic memory alone fails to infer habitual actions that extend beyond a single event. 
By contrast, semantic memory captures the repeated use of kitchen wet wipes, allowing the model to infer the correct answer through long-term reasoning.

\paragraph{Adaptive Retrieval on Multimodal Memory}
To further analyze how categories differ in their reliance on distinct memory modalities in \modelname{}, we quantify the usage proportion of each memory across all retrieval iterations per category. As shown in \cref{fig:memtype}, while episodic memory plays a foundational role across all tasks, certain categories tend to select it more frequently than other memory types: HabitInsight and RelationMap depend primarily on semantic memory, reflecting their reliance on reasoning over long-term patterns. In contrast, EntityLog and EventRecall benefit more from visual memory, which provides fine-grained perceptual details not fully captured by text. This selective utilization suggests that the model dynamically emphasizes the most relevant memory type for each category, leveraging the strength of each type in a context-dependent manner. These results confirm that different types of memory contribute distinct yet complementary strengths. 

\subsection{Dynamic Temporal Scope Retrieval}
\label{sec:dynamic_temporal_scope_retrieval}
We evaluate episodic memory retrieval performance across diverse temporal scales of events using temporal intersection over union (tIoU), which measures the overlap between retrieved and ground truth segments as the ratio of their intersection to their union duration. We compare \modelname{} with various models in temporal grounding, single-modality retrieval, long-form egocentric video retrieval, and keyframe selection. Details about baselines are given in \cref{sec:appendix_dynamic_temporal_scope_retrieval}. As shown in \cref{tab:dynamic_memory}, \modelname{} achieves significantly superior tIoU scores than strong baselines. Notably, reasoning-based retrieval and keyframe selection methods exhibit lower tIoU values, indicating difficulty in handling long input contexts. Moreover, \cref{fig:tiou} demonstrates that the superior tIoU is directly correlated with higher overall accuracy, particularly in understanding long videos.

\subsection{Efficacy of Multi-turn Retrieval}
We validate the effectiveness of our model's multi-turn approach by limiting the maximum number of retrieval steps. The results in \cref{fig:iter} show that performance consistently improves as the number of iterations increases across all benchmarks. Notably, on the EgoLifeQA benchmark, allowing a maximum of five steps yields a 9.3\% improvement over single-step retrieval. This gain arises because multiple iterations enable the retrieval agent to gather additional relevant information and refine its retrieval strategy when earlier attempts are suboptimal. An example of this refinement is shown in \cref{sec:appendix_refin}, where the model corrects an initially irrelevant retrieval to produce a more accurate and contextually grounded response.

\subsection{Analysis on Efficiency}
To assess the efficiency of our framework, we measure the end-to-end latency of \modelname{} on 100 randomly sampled queries from EgoLifeQA. As shown in \cref{fig:latency}, our method achieves a superior latency–accuracy trade-off compared with baselines. Long-video LLMs incur significantly higher inference latency, while still exhibiting relatively low performance. Although RAG- or memory-based approaches offer better latency, they often require substantial preprocessing and show a significant performance gap. In contrast, by allowing the retrieval agent to adaptively finish iterations and by retrieving only the relevant segments, \modelname{} achieves low latency and substantially higher accuracy.

\subsection{Efficacy of Memory Modules}
\label{sec:efficacy_of_memory_modules}
To enable effective long video reasoning, \modelname{} applies different strategies for each memory. \cref{tab:ablation_module} reports the results of \modelname{}-8B under various module configurations with details about each method in \cref{sec:appendix_efficacy_of_memory_modules}. For episodic memory, using a fixed single timescale or embeddings instead of graphs results in a 6.1\% and 4.4\% drop in average accuracy, respectively, highlighting the importance of multi-scale structured knowledge. For semantic memory, removing the consolidation process results in approximately a 7\% drop for the category that requires long-term reasoning, demonstrating the need for continuous integration of knowledge to support long-term reasoning. Finally, for visual memory, disabling its dual-mode retrieval leads to an accuracy drop of about 3\%, indicating that each mode contributes complementary benefits for retrieving particular scenes or accessing broader temporal ranges.

%% file: fig/fig_memtype.tex
\begin{figure}[t!]
    \centering
    \includegraphics[width=\linewidth]{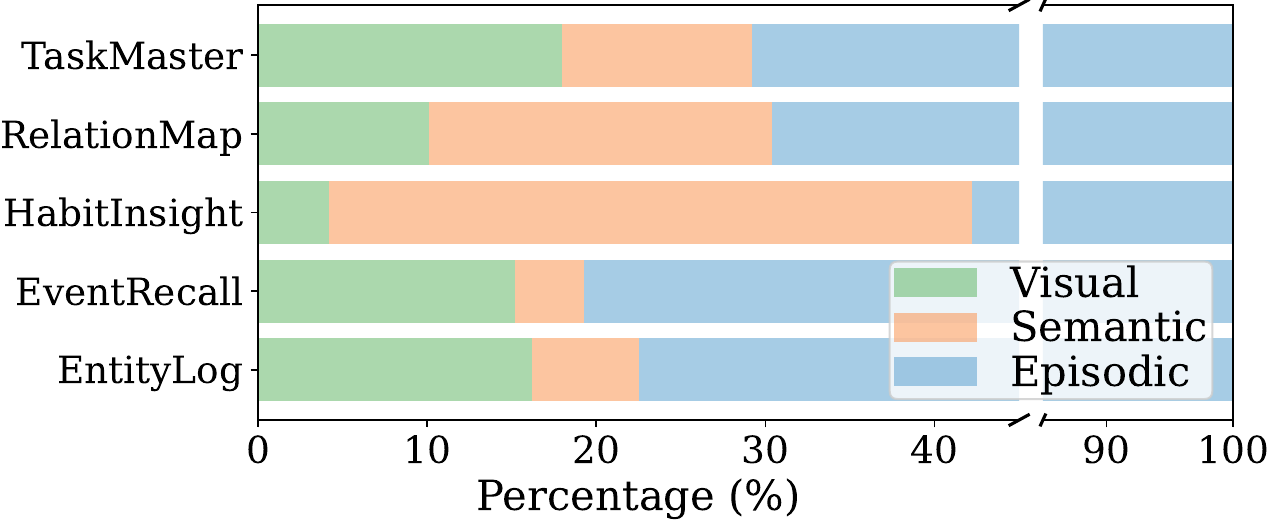}
    \vspace{-0.15in}
    \caption{Memory type utilization of \modelname{} on five distinctive categories in EgoLifeQA.} 
    \label{fig:memtype}
    \vspace{-0.1in}
\end{figure}

%% file: tab/tab_memory_type.tex
\begin{table*}[t]
\centering
\setlength{\tabcolsep}{3pt}
\renewcommand{\arraystretch}{1.1}
\caption{Performance of \modelname{} across multiple benchmarks using different memory types. E, S, and V denote episodic, semantic, and visual memories, respectively. Combinations with ``+'' indicate multiple memory types are used.}
\label{tab:memory_type}
\resizebox{\textwidth}{!}{
\begin{tabular}{lccccc>{\columncolor{mygray}}c ccccc>{\columncolor{mygray}}c cccc>{\columncolor{mygray}}c ccc}
\toprule
\multirow{2}{*}{\textbf{Model}} &
\multicolumn{6}{c}{\textbf{EgoLifeQA}} &
\multicolumn{6}{c}{\textbf{Ego-R1 Bench}} &
\multicolumn{5}{c}{\textbf{HippoVlog}} &
\multirow{2}{*}{\textbf{LVBench}} &
\multirow{2}{*}{\raisebox{-0.55\totalheight}{\makecell{\textbf{Video-MME} \\ \textbf{(L)}}}} &
\multirow{2}{*}{\textbf{Avg.}} \\
\cmidrule(lr){2-7} \cmidrule(lr){8-13} \cmidrule(lr){14-18}
 & Ent. & EvR. & Hab. & Rel. & Task & Avg. 
 & Ent. & EvR. & Hab. & Rel. & Task & Avg. 
 & Aud. & Vis. & A+V & Summ. & Avg. \\
\midrule
E & 57.6 & 61.1 & 70.5 & 61.6 & 69.8 & 62.6 & 54.5 & 70.7 & 53.9 & 52.6 & 57.9 & 57.0 & 72.4 & 73.2 & 68.4 & 80.4 & 73.6 & 60.6 & 72.7 & 64.9 \\
V & 40.8 & 35.7 & 36.1 & 34.4 & 39.7 & 37.2 & 36.5 & 34.1 & 23.1 & 31.6 & 28.2 & 34.2 & 35.2 & 66.4 & 54.8 & 48.8 & 51.3 & 47.4 & 64.2 & 44.9 \\
E+S & 56.8 & 61.9 & 73.8 & 62.4 & 71.4 & 63.4 & 59.3 & 68.3 & 69.2 & 57.9 & 60.5 & 61.0 & 70.8 & 75.2 & 68.8 & 80.4 & 73.8 & 58.8 & 74.1 & 66.8 \\
E+V & 59.2 & 63.5 & 70.5 & 60.8 & 68.8 & 63.3 & 65.1 & 68.3 & 53.9 & 47.4 & 57.9 & 63.0 & 73.2 & 77.2 & 70.8 & 79.6 & 75.2 & 59.8 & 76.0 & 66.9 \\
\rowcolor{myblue} \textbf{E+S+V} & \textbf{62.4} & \textbf{64.3} & \textbf{75.4} & \textbf{62.4} & \textbf{71.4} & \textbf{65.6} & \textbf{64.6} & \textbf{70.7} & \textbf{76.9} & \textbf{57.9} & \textbf{63.2} & \textbf{65.3} & \textbf{75.6} & \textbf{81.6} & \textbf{73.2} & \textbf{82.8} & \textbf{78.3} & \textbf{61.9} & \textbf{76.6} & \textbf{69.5} \\
\bottomrule
\end{tabular}
}
\end{table*}

%% file: fig/fig_qualitative.tex
\begin{figure*}[t!]
    \centering
    \includegraphics[width=\textwidth]{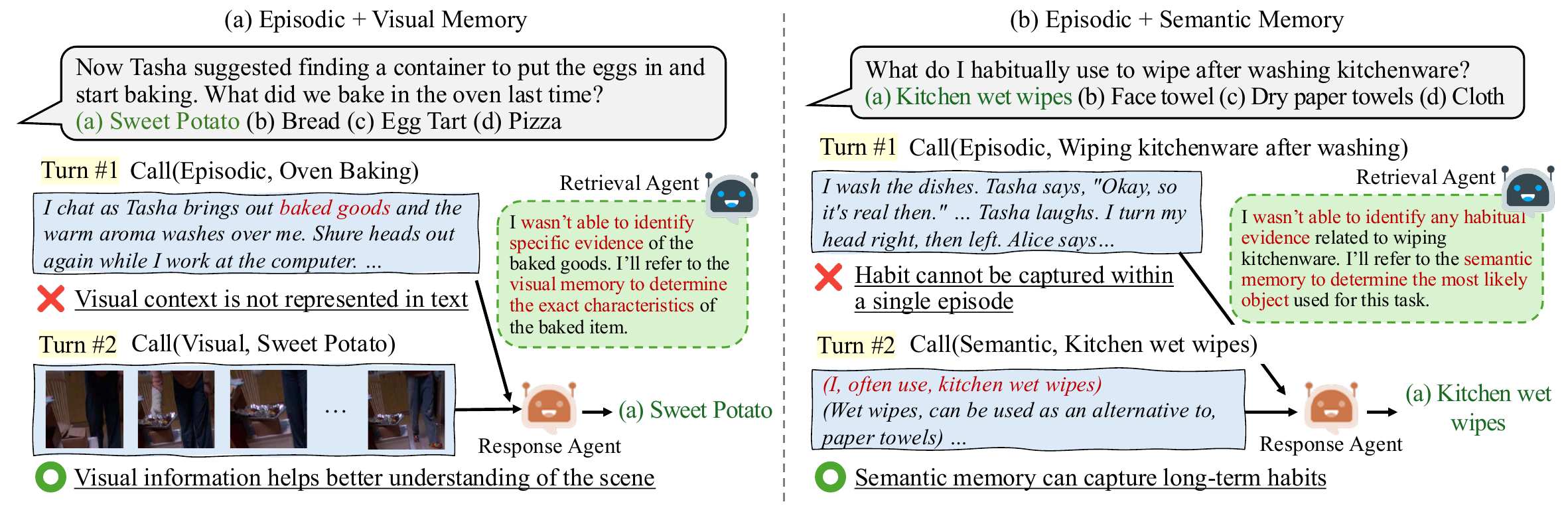}
    \vspace{-0.25in}
    \caption{\textbf{Qualitative results.} (a) Episodic memory alone cannot capture detailed visual context. The retrieval agent dynamically retrieves from visual memory, enabling access to fine-grained visual details. (b) To address the limitations of episodic memory in representing relationships or habitual behaviors, the retrieval agent proactively accesses semantic memory, allowing it to incorporate habitual knowledge.}
    \label{fig:qualitative}
    \vspace{-0.15in}
\end{figure*}

%% file: fig/fig_tables.tex
\begin{figure*}[t]
    \centering
    \begin{minipage}[t]{0.48\textwidth}
        \centering
        \setlength{\tabcolsep}{4.5pt}
        \renewcommand{\arraystretch}{1.09}
        \captionsetup{type=table}
        \caption{Average tIoU (\%) across three benchmarks.}
        \vspace{-0.05in}
        \label{tab:dynamic_memory}
        \resizebox{\linewidth}{!}{
        \begin{tabular}{lccc}
            \toprule
            \textbf{Model} & \textbf{EgoLifeQA} & \textbf{Ego-R1 Bench} & \textbf{LVBench} \\
            \midrule
            Time-R1~\citep{time-r1} & 0.58 & 0.59 & 2.70 \\
            \midrule
            Qwen3 Emb.~\citep{qwen3embedding} & 4.35 & 2.87 & 4.54 \\
            HippoRAG~\citep{hipporag} & 4.00 & 3.28 & 4.30 \\
            InternVideo2~\citep{internvideo2} & 3.36 & 2.60 & 3.55 \\
            \midrule
            EgoRAG~\citep{egolife} & 3.60 & 2.73 & 3.50 \\
            Ego-R1~\citep{ego-r1} & 3.70 & 2.89 & 3.60 \\
            \midrule
            AKS~\citep{aks} & 2.75 & 2.30 & 3.52 \\
            \midrule
            \rowcolor{myblue} \textbf{\modelname{} (Ours)} & \textbf{10.09} & \textbf{9.17} & \textbf{9.57} \\
            \bottomrule
        \end{tabular}
        }
    \end{minipage}
    \hfill
    \begin{minipage}[t]{0.48\textwidth}
    \centering
    \setlength{\tabcolsep}{1.7pt}
    \renewcommand{\arraystretch}{0.9}
    \captionsetup{type=table}
    \caption{Comparison of model variants by changing each module.}
    \vspace{-0.05in}
    \label{tab:ablation_module}
    \resizebox{\linewidth}{!}{
    \begin{tabular}{lccccc>{\columncolor{mygray}}c c}
        \toprule
        \multirow{2}{*}{\textbf{Model}} &
        \multicolumn{6}{c}{\textbf{EgoLifeQA}} &
        \textbf{LVBench} \\
        \cmidrule(lr){2-7} \cmidrule(lr){8-8}
         & Ent. & EvR. & Hab. & Rel. & Task & Avg. & Acc. \\
        \midrule
        \textbf{\textit{Episodic Memory}} \\
        Fixed Timescale & 44.8 & 51.6 & 60.7 & 51.2 & 58.7 & 51.8 & 47.9 \\
        Embedding Retrieval & 45.6 & 52.4 & 59.0 & 54.4 & 52.4 & 52.0 & 50.9 \\
        \midrule
        \textbf{\textit{Semantic Memory}} \\
        w/o Consolidation & 48.8 & 53.2 & 57.4 & 51.2 & \textbf{60.7} & 53.0 & 54.2 \\
        \midrule
        \textbf{\textit{Visual Memory}} \\
        Feature Retrieval & 45.6 & 51.6 & 62.3 & \textbf{58.4} & 55.6 & 53.6 & 52.4 \\
        Timestamp Retrieval & 41.6 & 50.0 & \textbf{63.9} & 56.8 & 54.0 & 51.8 & 52.9 \\
        \midrule
        \rowcolor{myblue} \textbf{\modelname{} (Ours)} & \textbf{49.6} & \textbf{56.4} & \textbf{63.9} & \textbf{58.4} & 58.7 & \textbf{56.4} & \textbf{55.4} \\
        \bottomrule
        \end{tabular}
    }
    \end{minipage}
\end{figure*}

%% file: fig/fig_figures.tex
\begin{figure*}[t]
    \centering
    \begin{minipage}[t]{0.31\textwidth}
        \centering
        \includegraphics[width=\linewidth]{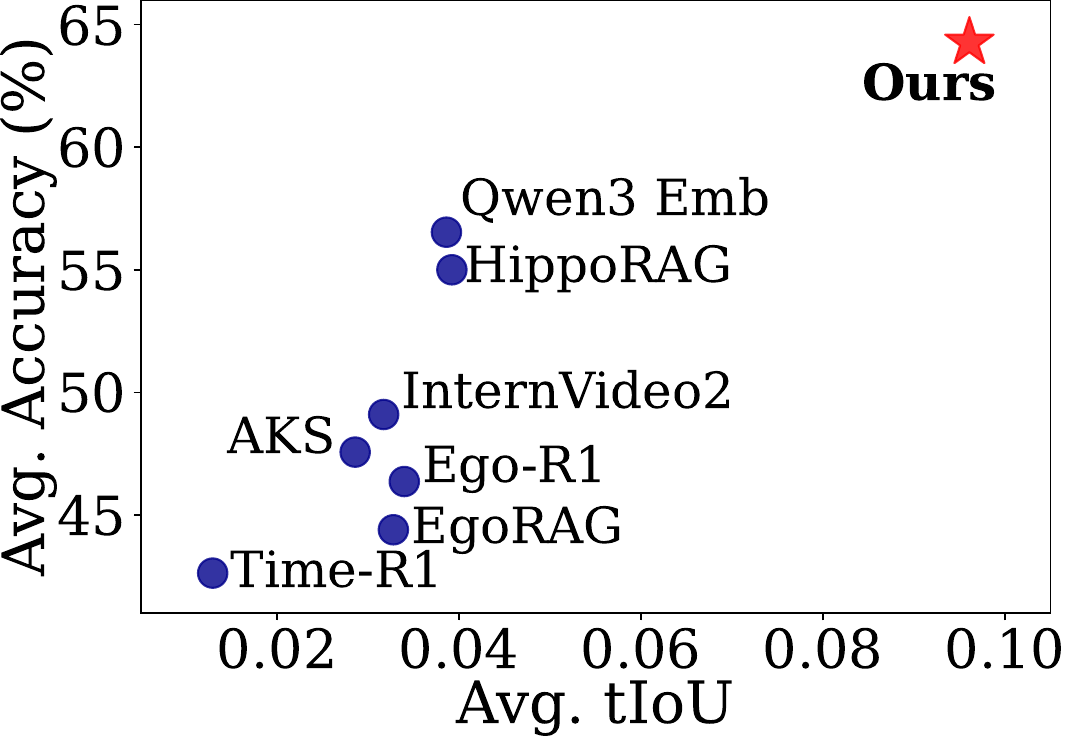}
        \vspace{-0.25in}
        \caption{Average tIoU and performance of \modelname{} and baselines.}
        \label{fig:tiou}
    \end{minipage}
    \hfill
    \begin{minipage}[t]{0.31\textwidth}
        \centering
        \includegraphics[width=\linewidth]{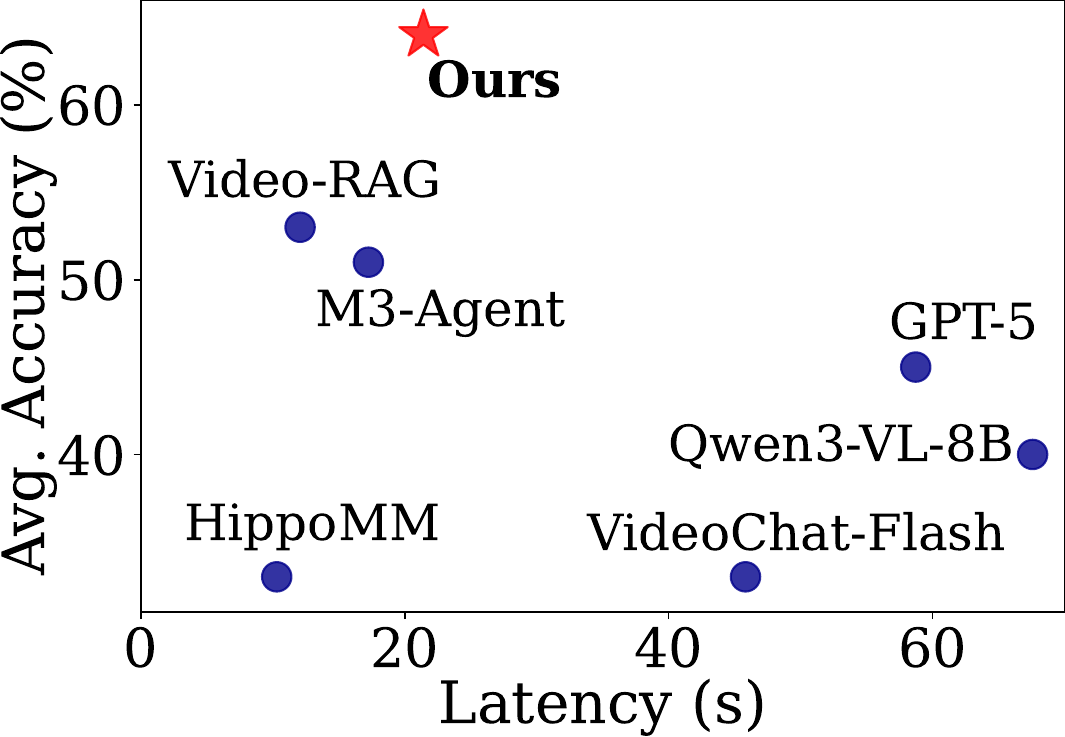}
        \vspace{-0.25in}
        \caption{Average latency and performance of \modelname{} and baselines.}
        \label{fig:latency}
    \end{minipage}
    \hfill
    \begin{minipage}[t]{0.31\textwidth}
        \centering
        \includegraphics[width=\linewidth]{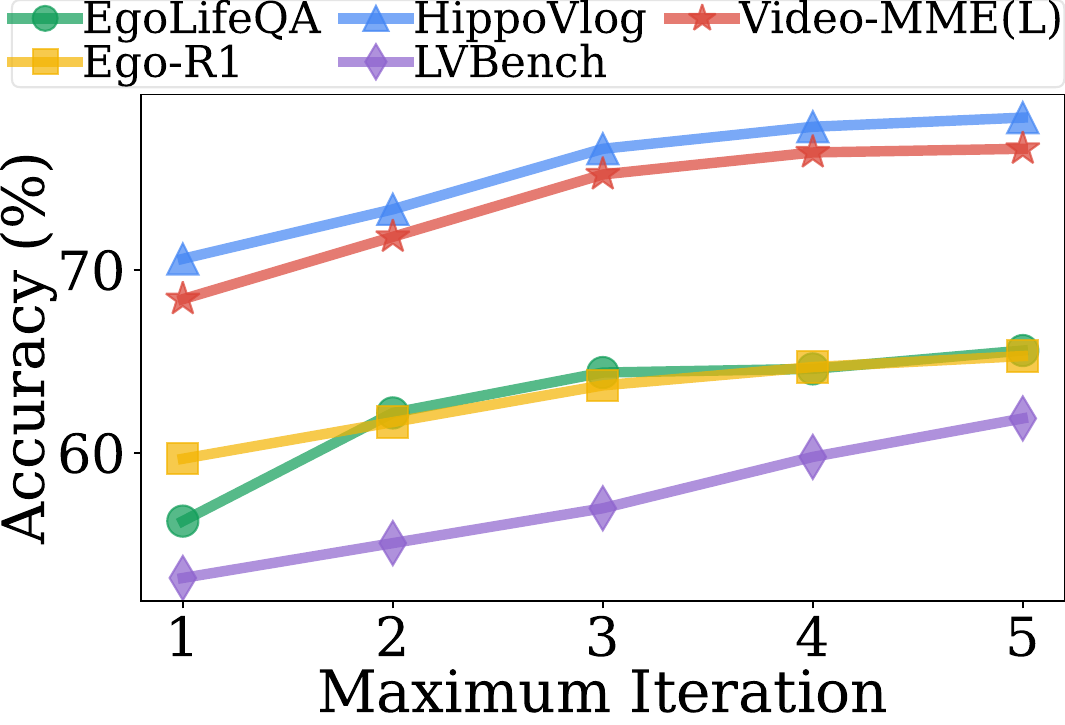}
        \vspace{-0.25in}
        \caption{Accuracy of \modelname{} with different maximum retrieval steps.}
        \label{fig:iter}
    \end{minipage}
\vspace{-0.1in}
\end{figure*}

%% file: sec/6_conclusion.tex
\section{Conclusion}
\label{sec:conclusion}

We propose \modelname{}, a novel memory agent designed to perceive and remember the world as represented in long video streams. To address the challenges of long video reasoning, we introduce a multimodal, multi-scale memory that integrates textual and visual information through adaptive retrieval. By constructing separate memories across different modalities and timescales, together with a retrieval agent that iteratively identifies relevant information, our approach enables effective and flexible reasoning over long videos. We validate our model on multiple benchmarks ranging from hour- to week-long videos, demonstrating that \modelname{} provides a promising solution capable of robust performance across various long video reasoning tasks.

%% file: sec/X_acknowledgement.tex
\section*{Acknowledgements}

This work was supported by the Institute for Information \& communications Technology Technology Planning \& Evaluation (IITP) grant funded by the Korea government (MSIT) (RS-2019-II190075, Artificial Intelligence Graduate School Program (KAIST)), the National Research Foundation of Korea (NRF) grant funded by the Korea government (MSIT) (RS-2023-00256259), the Korea Machine Learning Ledger Orchestration for Drug Discovery Project (K-MELLODDY) grant funded by the Ministry of Health \& Welfare and Ministry of Science and ICT, Republic of Korea (RS-2024-00460870), the Institute of Information \& communications Technology Planning \& Evaluation (IITP) grant funded by the Ministry of Science and ICT (MSIT) of the Republic of Korea in connection with the Global AI Frontier Lab International Collaborative Research (RS-2024-00469482 \& RS-2024-00509279), the Institute of Information \& communications Technology Planning \& Evaluation (IITP) grant funded by the Korea government (MSIT) (RS-2022-II220713, Meta-learning Applicable to Real-world Problems), and the NTU Start Up grant.

%% file: sec/X_suppl.tex
\clearpage
\maketitlesupplementary

\newcounter{sectiondisp}

\let\oldsection\section
\renewcommand{\section}[1]{
    \stepcounter{sectiondisp}
    \oldsection{#1}
}

\renewcommand{\thesection}{\Alph{sectiondisp}}




In this supplementary material, we provide additional details on the dataset (\cref{sec:appendix_dataset}), additional implementation details (\cref{sec:appendix_implementation_details}) and descriptions on experiments (\cref{sec:appendix_additional_desc_exps}). We also present detailed and additional experimental results (\cref{sec:appendix_experimental_results,sec:appendix_additional_exp_results}), qualitative analyses (\cref{sec:appendix_qualitative}), and a discussion of limitations and broader impacts (\cref{sec:appendix_limitation}).

\section{Additional Details on Dataset}
\label{sec:appendix_dataset}

In this section, we provide additional details for each dataset used in our experiments. \cref{tab:appendix_dataset} summarizes the datasets, including the number of queries, domain categories, and the average video duration.

\input{tab/tab_appendix_dataset}

\subsection{EgoLifeQA} EgoLifeQA~\citep{egolife} is a set of questions designed to test the capability of models to understand and remember everyday life from week-long video recordings. It includes questions that require recalling past events, tracking object locations, and reasoning over long-term activities. In our experiments, we use questions from the perspective of a single participant (A1: JAKE), along with his corresponding video stream, which spans 44.3 hours. The benchmark is organized into five distinct categories as follows.

\paragraph{EntityLog (Ent.)} Questions that require recalling information about objects, such as their locations, states, or interactions. (Example: ``Who used the screwdriver first?")
\paragraph{EventRecall (EvR.)} Questions that ask about specific past events, including what happened, when it occurred, and relevant context. (Example: ``Shure mentioned Tiramisu, when was the last time we discussed making Tiramisu?")
\paragraph{HabitInsight (Hab.)} Questions aimed at identifying a person's recurring behaviors or long-term activity patterns. (Example: ``What food does Alice love to eat?")
\paragraph{RelationMap (Rel.)} Questions involving understanding social relationships and interactions between people. (Example: ``Who usually sings when Shure plays the guitar?")
\paragraph{TaskMaster (Task)} Questions focused on ongoing or pending tasks that require reasoning about what actions still need to be completed. (Example: ``What are we planning to do in the afternoon?")

\subsection{Ego-R1 Bench} Ego-R1 Bench~\citep{ego-r1} is designed as a complementary evaluation to EgoLifeQA, but with a distinct focus on model reasoning. While both benchmarks focus on the same week-long egocentric video, Ego-R1 Bench targets multi-step, tool-augmented reasoning over ultra-long video. We reorganize query types of Ego-R1 Bench to the category adopted by EgoLifeQA, as shown in \cref{tab:egor1_cate}.

\begin{table}[h]
    \centering
    \caption{Classification of queries under the EgoLifeQA category.}
    \vspace{-0.05in}
    \label{tab:egor1_cate}
    \resizebox{\linewidth}{!}{
    \begin{tabular}{ll}
        \toprule
        \textbf{Category} & \textbf{Ego-R1 Category} \\
        \midrule
        EntityLog & EntityLog, FoodLog, HealthLog, TechLog \\
        EventRecall & EventRecall, Event Recollection, Event Memory \\
        HabitInsight & HabitInsight, Behavior Habit(s) \\
        RelationMap & RelationMap, Interpersonal Relationships \\
        TaskMaster & TaskMaster, Future Plan(s) \\
        \bottomrule
    \end{tabular}
    }
\end{table}

\subsection{HippoVlog} HippoVlog~\citep{hippomm} contains 25 daily vlog videos with 1,000 multiple-choice questions for continuous audiovisual event understanding. The benchmark evaluates a model’s ability to handle modality-specific information, with \textbf{Auditory (Aud.)} questions requiring reasoning over the audio stream (or transcript) and \textbf{Visual (Vis.)} questions focusing on the visual content. \textbf{Auditory+Visual (A+V)} queries test the model’s ability to integrate information across both modalities, while \textbf{Summarization (Summ.)} questions assess higher-level reasoning over long temporal spans, requiring synthesis of events and semantic understanding from the continuous video.

\subsection{LVBench} LVBench~\citep{lvbench} consists of 103 long videos, typically longer than an hour, with 1,549 multiple-choice questions for extreme long video understanding. The videos cover a general and diverse set of domains. Questions include both visual perception for recognizing entities or events in short segments and summarization for higher-level reasoning across extended sequences, evaluating models’ ability to integrate information over both local and long-horizon contexts. In our experiments, we categorize questions into three groups based on their segment length, defined as the duration of video required to answer the question: \textbf{Short} ($<$30s), \textbf{Medium (Med.)} (30s$\sim$5min), and \textbf{Long} ($>$5min). We excluded 15 questions without segment tags, leaving 1,534 questions in total for evaluation.

\subsection{Video-MME} Video-MME~\citep{videomme} is a comprehensive video understanding benchmark with 2,700 questions and varying video durations. In this experiment, we use only the long subset ($>$30min), containing 900 questions, to assess the model’s capability on long video reasoning. We adopt the categories provided by the benchmark, with acronyms as follows: Action Reasoning (ARES), Action Recognition (AREC), Attribute Perception (ATTR), Counting Problem (CNT), Information Synopsis (ISYN), OCR Problems (OCR), Object Reasoning (ORES), Object Recognition (OREC), Spatial Perception (SPER), Spatial Reasoning (SRES), Temporal Perception (TPER), and Temporal Reasoning (TRES).

\section{Additional Implementation Details}
\label{sec:appendix_implementation_details}

We provide additional details on the baseline setup (\cref{sec:appendix_baselines}), the configuration of our proposed \modelname{} (\cref{sec:appendix_configurations}), and the prompts used (\cref{sec:appendix_used_prompts}).

\subsection{Baseline Setup}
\label{sec:appendix_baselines}

\paragraph{Base Models \& Long Video LLMs}
For all base models and long video LLMs, the video input is uniformly sampled at 0.5 fps and capped at 768 frames due to context length limitations, as described in \cref{sec:intro}. In this setting, the models operate solely on visual frames without access to video captions or speech transcripts.

\paragraph{RAG-based Video LLMs}
For text-based RAG video models, we construct a knowledge base from video captions. Specifically, each video is segmented into 30 second chunks, and set of captions from these segments serve as retrieval pool. LightRAG~\citep{lightrag} performs dual-level retrieval, selecting either fine-grained (low-level) or abstracted (high-level) information from the knowledge graph generated from set of captions depending on the query. HippoRAG~\citep{hipporag}, in contrast, retrieves raw captions ranked by their PPR scores, treating each caption as a separate document.
For Video-RAG~\citep{video-rag} model, retrieval is performed directly on the raw video using tools such as optical character recognition (OCR) and automatic speech recognition (ASR) to extract textual signals. Unless otherwise stated, we follow the retrieval specifications described in each model’s corresponding paper or implementation.

\paragraph{Memory-based Video LLMs}
Memory-based video LLMs construct explicit memories from the video stream. For EgoRAG~\citep{egolife} and Ego-R1~\citep{ego-r1}, which build hierarchical textual memories, we use the same temporal granularity applied when constructing \modelname{}’s memory. For models that perform iterative reasoning, including Ego-R1~\citep{ego-r1} and M3-Agent~\citep{m3-agent}, we evaluate the checkpoints released by authors and set the maximum number of reasoning iterations to 5 to ensure consistent evaluation across all systems. All other implementation details follow the official specifications provided by the respective authors.

\subsection{\modelname{}}
\label{sec:appendix_configurations}
To construct multi-scale episodic memory, video captioning is performed at each temporal unit by passing sampled video frames along with transcripts generated using Distil-Whisper large-v3.5~\citep{distil-whisper}. Moreover, we tailor the temporal resolutions to each dataset’s duration. For EgoLifeQA and Ego-R1 Bench, which contain week-long videos, we use four broad timescales: 30 seconds, 3 minutes, 10 minutes, and 1 hour. For HippoVlog, LVBench, and Video-MME, which contain shorter recordings averaging about an hour, we adopt shorter timescales of 10 seconds, 30 seconds, 3 minutes, and 10 minutes to better match their temporal structure. For semantic memory, triplets with a similarity score above 0.6 are consolidated using an LLM, and the top 10 triplets are retrieved at query time. The retrieval agent is limited to a maximum of five iterations, consistent with the baseline evaluation setting.

\subsection{Prompts}
\label{sec:appendix_used_prompts}
To construct and retrieve memory, and to generate the final response of \modelname{}, we employ carefully optimized prompts for use with an LLM. In particular, we use prompts for video captioning~(\cref{fig:prompt_video_caption}), episodic triple extraction~(\cref{fig:prompt_episodic_ner_extraction,fig:prompt_episodic_triple_extraction}) and multi-scale memory construction~(\cref{fig:prompt_episodic_multi_construction}), adapted from \citet{egolife}. Furthermore, we utilize prompts for multi-scale memory retrieval~(\cref{fig:prompt_episodic_multi_retrieval}), semantic triple extraction~(\cref{fig:prompt_semantic_triple_extraction}), semantic consolidation~(\cref{fig:prompt_semantic_triple_consolidation}), iterative reasoning by the retrieval agent~(\cref{fig:prompt_retrieval_agent}), and final response generation~(\cref{fig:prompt_response_agent}).

\input{tab/tab_main_full}

\section{Additional Description on Experiments}
\label{sec:appendix_additional_desc_exps}
In this section, we provide additional description of the settings used in our ablation experiments.

\subsection{Dynamic Temporal Scope Retrieval~(\cref{sec:dynamic_temporal_scope_retrieval})}
\label{sec:appendix_dynamic_temporal_scope_retrieval}
To evaluate performance on dynamic temporal reasoning with \modelname{}, we employ several approaches, including temporal grounding model, embedding-based retrieval models, hierarchical retrieval models, and keyframe selection method. For each method, we measure tIoU using either the returned timestamps or the timestamps of the selected content.
For the temporal grounding model, we use Time-R1~\citep{time-r1}, with a slightly modified prompt that enables it to return both the evidence timestamps and the corresponding grounded responses. We sample videos at 0.5 fps and provide up to 768 frames. For embedding-based and hierarchical retrieval models, we follow the configurations described in \cref{sec:appendix_baselines}. Additionally, we include Qwen3 Emb., which applies the Qwen3-Embedding-4B~\citep{qwen3embedding} text encoder for caption retrieval, and InternVideo2, which encodes each segment using InternVideo2~\citep{internvideo2} as an video encoder with uniform 16 frame averaging to enable segment-level retrieval. Both methods retrieve 30 second segments based on similarity search. For key frame selection, we apply AKS~\citep{aks}, which selects keyframes from the 0.5 fps sampled sequence. For tIoU evaluation, we interpret frames as representing their corresponding 30 second segments.

\subsection{Efficacy of Memory Modules~(\cref{sec:efficacy_of_memory_modules})}
\label{sec:appendix_efficacy_of_memory_modules}
To assess the contribution of each component within \modelname{}’s multimodal memory system, we evaluate several ablated variants in \cref{sec:efficacy_of_memory_modules}. In this section, we detail each variant of \modelname{} created by selectively disabling a specific component. For episodic memory variants, we first construct a \textbf{fixed timescale} variant by replacing hierarchical episodic memory with a single fixed timescale memory. Specifically, we use the episodic memory of the finest granularity timescale. We also experiment an \textbf{embedding retrieval} variant in which the model's graph-based episodic retrieval is replaced with an embedding-based similarity search using Qwen-Embedding-4B. To examine the effect of semantic consolidation, we use a \textbf{w/o consolidation} version that bypasses the consolidation procedure to update the memory and instead store the raw extracted triplets without any update to existing memory. Finally, for visual memory, we ablate components of dual-retrieval mechanism by evaluating systems that rely exclusively on either \textbf{feature retrieval} through natural-language keyword search or \textbf{timestamp retrieval} based purely on temporal indices.

\section{Detailed Experimental Results}
\label{sec:appendix_experimental_results}

In this section, we provide extended results and analyses of the experimental results.

\input{tab/tab_main_videomme}
\input{fig/fig_appendix_memtype_hippovlog}

\paragraph{Main results} \cref{tab:main_results_full,tab:main_results_videomme} present the category-wise performance breakdown of \modelname{} and baseline methods. Beyond overall benchmark averages, \modelname{} consistently outperforms existing approaches across most categories. Notably, the gains are particularly pronounced in categories that rely on visual information. For instance, in the EntityRecall category of EgoLifeQA, where visual cues can help answering, \modelname{} exceeds the previous best method, Ego-R1, by a substantial 11.2\%. Similarly, on HippoVlog, our model achieves a 4\% improvement in the Aud. and A+V categories, both of which require visual reasoning. These margins are greater than those observed in categories that do not explicitly depend on visual content, highlighting the strong advantage of our multimodal multi-memory architecture.

\paragraph{Efficacy of multimodal memory} \cref{fig:memtype_hippovlog} shows memory type utilization of our model on HippoVlog benchmark, where categories are grouped by their modality requirements. The Audio category requires reasoning over spoken content and therefore is expected to depend primarily on textual memory derived from caption transcripts, while the Visual category focuses on visual understanding and correspondingly is designed to rely more on visual memory. Our results clearly support these expectations, showing that the Audio category predominantly activates textual memory while the Visual category relies heavily on visual memory, indicating that each category effectively leverages the required memory. Moreover, the Summarization category, which requires long-term reasoning, utilizes semantic memory more than any other category, demonstrating the complementary roles and effectiveness of each memory module in handling different reasoning demands. Together with this distribution of memory usage and the demonstrated performance gains in \cref{tab:memory_type}, these underscore the effectiveness of our multimodal multi-memory framework.

\paragraph{Dynamic temporal scope retrieval} \cref{tab:appendix_tiou,tab:appendix_dyna_perf} detail the per-category tIoU and accuracy results for \modelname{} and baseline methods. While \modelname{} significantly outperforms existing baselines on average, the results on LVBench particularly highlight the effectiveness of our dynamic episodic memory. In LVBench’s Long category, where answering requires reasoning over more than five minutes of video, \modelname{} outperforms the baselines by a notably larger margin than in categories that require shorter timescale, underscoring its ability to flexibly retrieve and integrate information over diverse temporal spans.

\clearpage
\input{tab/tab_dynamic_memory}
\clearpage

\section{Additional Experimental Results}
\label{sec:appendix_additional_exp_results}
We additionally present experimental results supporting the design of \modelname{}, including ablation studies on backbone configurations (\cref{sec:appendix_exp_backbone}) and the impact of temporal scales (\cref{sec:appendix_exp_temporal}).

\subsection{Generalization to Different Backbones}
\label{sec:appendix_exp_backbone}

To evaluate the flexibility and robustness of \modelname{} across different backbone models, we conduct experiments with a diverse set of configurations. Specifically, in addition to the setup based on the GPT-5 model series and VLM2Vec-V2, we further incorporate Gemini 3 Flash~\citep{gemini3} and Qwen3-VL-Embedding-2B~\citep{qwen3vlembedding}. As shown in \cref{tab:diff_backbones}, \modelname{} demonstrates strong robustness to backbone selection, with the Gemini-based variant even outperforming others on EgoLifeQA. These results highlight that \modelname{} generalizes well across different backbone architectures and can be seamlessly integrated with a wide range of state-of-the-art models without requiring architecture-specific modifications.

\input{tab/tab_backbones}

\subsection{Impact of Temporal Scales}
\label{sec:appendix_exp_temporal}

While multiscale episodic memory improves overall performance, we verify that these gains result from the multiscale architecture rather than specific temporal constraints. The temporal scales used in our experiments are chosen based on empirical statistics of real-world event durations. To assess the sensitivity of \modelname{} to these specific values, we introduce perturbations to the temporal scales and report the performance on EgoLifeQA in \cref{tab:temp_scale}. The results demonstrate that \modelname{} maintains consistent performance across these variations, indicating that the improvements stem from the multiscale memory design itself, rather than a reliance on precisely calibrated temporal windows.

\input{tab/tab_temp_scale}

\section{Qualitative Results}
\label{sec:appendix_qualitative}
In this section, we qualitatively analyze \modelname{}'s memory construction (\cref{sec:appendix_qual_mem_const}) and its multi-turn reasoning and refinement capabilities (\cref{sec:appendix_refin}).

\subsection{Memory Construction}
\label{sec:appendix_qual_mem_const}
\cref{tab:episodic_triple_extraction} presents an example of episodic triplet extraction. Given a caption generated from sampled frames of a segment along with its corresponding transcript, an LLM is prompted (using the prompt in \cref{fig:prompt_episodic_triple_extraction}) to extract episodic triplets. Semantic triplets are extracted using a different prompt (\cref{fig:prompt_semantic_triple_extraction}), designed to focus on long-term dependencies and capture more abstract relationships across the segments, as shown in \cref{tab:semantic_triple_extraction}. To better capture persistent knowledge across segments, we introduce semantic consolidation, which incrementally updates the semantic graph by integrating new triplets and resolving conflicts. Using embedding-based matching and an LLM, duplicated or conflicting triplets are removed, and new or revised ones are added, generating an evolving semantic memory, as shown in \cref{tab:semantic_triple_consolidation}. For instance, the new triplet ``[I, uses WeChat for, money transfers]" is merged with the existing triplet to consolidate redundant information, and conflicting triplets, such as ``[Lucia, dislikes, overly sweet food]" versus ``[Lucia, likes, sweet desserts]", are removed to ensure consistency in the semantic memory.

\subsection{Multi-turn Refinement}
\label{sec:appendix_refin}
\modelname{} demonstrates the effectiveness of multi-turn reasoning by progressively refining its retrieval strategy to answer questions, as shown in \cref{tab:qualitative_multi_turn_refinement}. In this example, the first round retrieves episodic memory using a narrow keyword focused on the ``discussion" of the air conditioning, but it provides insufficient detail about the activity. In the second round, the model expands to a more general keyword, ``air conditioning", which enables retrieval of every scene where the air conditioning is involved to obtain sufficient textual evidence. Moreover, in the third round, since the textual evidence fails to capture specific visual details of the scene, \modelname{} refines its strategy to retrieve video frames corresponding to the relevant timestamp. Through this stepwise process, \modelname{} effectively refines its search strategy with different keyword strategies and memory types to respond to the question.

\section{Limitation and Broader Impact}
\label{sec:appendix_limitation}

While \modelname{} serves as an effective multimodal memory agent for long video reasoning, it still requires careful preprocessing, including video captioning, triplet extraction, and semantic consolidation. Yet, this limitation is not unique to our approach but a broader constraint shared by existing memory-based video LLMs. For example, M3-Agent~\citep{m3-agent} incurs even heavier preprocessing due to its reliance on entity recognition, and many other approaches operate with offline preprocessing. In contrast, \modelname{} is designed for online operation. Memories are updated at fixed intervals (e.g., every 10 seconds), and the required preprocessing for each segment can be performed within these windows. Moreover, new information can be seamlessly integrated into the knowledge graph, and our consolidation mechanism efficiently refines the knowledge base without requiring the reconstruction of memory from scratch.

With strong long-term reasoning capabilities and support for real-time updates, \modelname{} serves as a practical solution for streaming scenarios such as egocentric assistants and embodied agents. This foundation enables richer and more persistent assistance for everyday tasks and accessibility. However, the continuous accumulation of structured knowledge over periods of time raises serious privacy and security concerns. Real-world deployments must therefore enforce safeguard policies, including strict access controls, secure data handling, and privacy protections.

\input{tab/qualitative/tab_episodic_triple_extraction}
\clearpage
\input{tab/qualitative/tab_semantic_triple_extraction}
\input{tab/qualitative/tab_semantic_triple_consolidation}
\input{tab/qualitative/tab_multi_turn_refinement}

\input{fig/prompts/fig_prompt_video_caption}
\input{fig/prompts/fig_prompt_episodic_ner_extraction}
\input{fig/prompts/fig_prompt_episodic_triple_extraction}
\input{fig/prompts/fig_prompt_episodic_multi_construction}
\input{fig/prompts/fig_prompt_episodic_multi_retrieval}
\input{fig/prompts/fig_prompt_semantic_triple_extraction}
\input{fig/prompts/fig_prompt_semantic_consolidation}
\input{fig/prompts/fig_prompt_retrieval_agent}
\input{fig/prompts/fig_prompt_response_agent}

%% file: tab/tab_appendix_dataset.tex
\begin{table}[h]
    \centering
    \caption{Summary of benchmark datasets used in experiments.}
    \vspace{-0.05in}
    \resizebox{\linewidth}{!}{
        \begin{tabular}{l c c c c}
            \toprule
            \textbf{Dataset} & \textbf{\# Queries} & \textbf{Domain} & \textbf{Avg. Video Length} \\
            \midrule
            EgoLifeQA~\citep{egolife} & 500 & Egocentric & 44.3h \\
            Ego-R1 Bench~\citep{ego-r1} & 300 & Egocentric & 44.3h \\
            HippoVlog~\citep{hippomm} & 1,000 & Vlog & 0.45h \\
            LVBench~\citep{lvbench} & 1,534 & General & 1.14h \\
            Video-MME (L)~\citep{videomme} & 900 & General & 0.69h \\
            \bottomrule
        \end{tabular}
    }
    \label{tab:appendix_dataset}
\end{table}

%% file: tab/tab_main_full.tex
\begin{table*}[t]
\centering
\setlength{\tabcolsep}{3pt}
\caption{Category-wise performance breakdown of \modelname{} and baselines on EgoLifeQA, Ego-R1 Bench, HippoVlog, and LVBench.}
\label{tab:main_results_full}
\resizebox{\textwidth}{!}{
\begin{tabular}{lccccc>{\columncolor{mygray}}c ccccc>{\columncolor{mygray}}c cccc>{\columncolor{mygray}}c ccc>{\columncolor{mygray}}c}
\toprule
\multirow{2}{*}{\textbf{Model}} &
\multicolumn{6}{c}{\textbf{EgoLifeQA}} &
\multicolumn{6}{c}{\textbf{Ego-R1 Bench}} &
\multicolumn{5}{c}{\textbf{HippoVlog}} &
\multicolumn{4}{c}{\textbf{LVBench}} \\
\cmidrule(lr){2-7} \cmidrule(lr){8-13} \cmidrule(lr){14-18} \cmidrule(lr){19-22}
 & Ent. & EvR. & Hab. & Rel. & Task & Avg.
 & Ent. & EvR. & Hab. & Rel. & Task & Avg.
 & Aud. & Vis. & A+V & Summ. & Avg.
 & Short & Med. & Long & Avg. \\
\midrule
\multicolumn{22}{l}{\textbf{\textit{Base Models}}} \\
Qwen3-VL-8B~\citep{qwen3vl} 
 & 35.2 & 30.2 & 39.3 & 46.4 & 46.0 & 38.6
 & 31.8 & 41.5 & 38.5 & 42.1 & 44.7 & 35.7
 & 73.6 & 74.0 & 69.2 & 80.8 & 74.4
 & 48.8 & 44.4 & 53.4 & 48.3 \\
Gemini 2.5 Pro~\citep{gemini2.5}
 & 43.2 & 40.5 & 41.0 & 55.2 & 52.4 & 46.4
 & 43.9 & 56.1 & 53.9 & 47.4 & 47.4 & 46.7
 & 69.2 & 75.2 & 63.6 & 80.0 & 72.0
 & 57.1 & 52.2 & 65.2 & 57.0 \\
GPT-5~\citep{gpt5}
 & 47.2 & 42.1 & 47.5 & 53.6 & 55.6 & 48.6
 & 41.8 & 58.5 & 53.9 & 52.6 & 50.0 & 46.3
 & 73.6 & 75.6 & 69.2 & \textbf{84.4} & 75.7
 & \textbf{59.1} & 59.1 & 69.1 & 60.4 \\
\midrule
\multicolumn{22}{l}{\textbf{\textit{Long Video LLMs}}} \\
VideoChat-Flash~\citep{videochat-flash}
 & 28.8 & 32.5 & 37.7 & 37.6 & 38.1 & 34.2
 & 43.4 & 43.9 & 38.5 & 31.6 & 44.7 & 42.7
 & 60.8 & 59.2 & 56.4 & 55.6 & 58.0
 & 34.9 & 23.1 & 44.6 & 33.2 \\
Time-R1~\citep{time-r1}
 & 39.2 & 50.8 & 65.6 & 48.8 & 47.6 & 48.8
 & 49.2 & 48.8 & 46.2 & 42.1 & 44.7 & 48.0
 & 58.2 & 58.2 & 49.4 & 52.4 & 54.6
 & 32.1 & 23.6 & 40.2 & 31.1 \\
Video-RTS~\citep{video-rts}
 & 40.8 & 48.4 & 62.3 & 48.8 & 47.6 & 48.2
 & 47.6 & 46.3 & 53.9 & 52.6 & 47.4 & 48.0
 & 58.8 & 62.0 & 56.8 & 58.4 & 59.0
 & 43.4 & 25.7 & 49.5 & 39.8 \\
\midrule
\multicolumn{22}{l}{\textbf{\textit{RAG-based Video LLMs}}} \\
LightRAG~\citep{lightrag}
 & 40.8 & 48.4 & 67.2 & 50.4 & 44.4 & 48.8
 & 54.0 & 61.0 & 46.2 & 42.1 & 42.1 & 52.3
 & 51.6 & 46.0 & 44.8 & 47.2 & 47.4
 & 30.2 & 28.6 & 34.3 & 30.4 \\
HippoRAG~\citep{hipporag}
 & 48.8 & 60.3 & 70.5 & 60.8 & 66.7 & 59.6
 & 54.5 & 65.9 & 69.2 & 52.6 & 50.0 & 56.0
 & 72.4 & 53.2 & 54.0 & 73.2 & 63.2
 & 54.9 & 47.5 & 62.3 & 54.0 \\
Video-RAG~\citep{video-rag}
 & 49.6 & 56.3 & 67.2 & 55.2 & 54.0 & 55.4
 & 48.7 & 58.5 & 53.9 & 47.4 & 44.7 & 49.7
 & 63.2 & 64.8 & 63.6 & 68.8 & 65.1
 & 32.9 & 30.2 & 39.7 & 33.1 \\
\midrule
\multicolumn{22}{l}{\textbf{\textit{Memory-based Video LLMs}}} \\
EgoRAG~\citep{egolife}
 & 40.0 & 56.3 & 62.3 & 54.4 & 52.4 & 52.0
 & 46.6 & 56.1 & 46.2 & 47.4 & 55.3 & 49.0
 & 64.8 & 53.2 & 47.6 & 64.4 & 57.5
 & 32.4 & 32.0 & 31.9 & 32.2 \\
Ego-R1~\citep{ego-r1}
 & 51.2 & 53.2 & 63.9 & 50.4 & 50.8 & 53.0
 & 50.8 & 63.4 & 38.5 & 36.8 & 57.9 & 52.0
 & 57.2 & 58.8 & 52.0 & 67.2 & 58.8
 & 32.5 & 36.5 & 37.3 & 34.1 \\
HippoMM~\citep{hippomm}
 & 45.6 & 53.2 & 70.5 & 55.2 & 58.7 & 54.6
 & 51.9 & 56.1 & 46.2 & 52.6 & 57.9 & 53.0
 & 68.8 & 77.6 & 59.2 & 82.0 & 71.9
 & 40.7 & 33.3 & 35.8 & 38.2 \\
M3-Agent~\citep{m3-agent}
 & 44.4 & 54.8 & 62.3 & 56.8 & 54.0 & 53.5
 & 52.4 & 58.5 & 38.5 & 42.1 & 52.6 & 52.0
 & 68.4 & 72.4 & 50.8 & 70.4 & 65.5
 & 53.0 & 40.7 & 48.5 & 49.3 \\
\midrule
\rowcolor{myblue} \multicolumn{22}{l}{\textbf{\textit{\modelname{} (Ours)}}} \\
\rowcolor{myblue} \textbf{\modelname{}-8B}
 & 49.6 & 56.4 & 63.9 & 58.4 & 58.7 & 56.4
 & 48.2 & 63.4 & 53.9 & 52.6 & 57.9 & 52.0
 & 69.6 & 73.6 & 65.2 & 70.4 & 69.7
 & 55.0 & 54.1 & 59.8 & 55.4 \\
\rowcolor{myblue} \textbf{\modelname{}-GPT}
 & \textbf{62.4} & \textbf{64.3} & \textbf{75.4} & \textbf{62.4} & \textbf{71.4} & \textbf{65.6}
 & \textbf{64.6} & \textbf{70.7} & \textbf{76.9} & \textbf{57.9} & \textbf{63.2} & \textbf{65.3}
 & \textbf{75.6} & \textbf{81.6} & \textbf{73.2} & 82.8 & \textbf{78.3}
 & 58.3 & \textbf{65.4} & \textbf{72.1} & \textbf{61.9} \\
\bottomrule
\end{tabular}
}
\end{table*}

%% file: tab/tab_main_videomme.tex
\begin{table*}[t]
\centering
\setlength{\tabcolsep}{3pt}
\renewcommand{\arraystretch}{1.1}
\caption{Category-wise performance breakdown of \modelname{} and baselines on Video-MME (L).}
\label{tab:main_results_videomme}
\resizebox{0.8\textwidth}{!}{
\begin{tabular}{lcccccccccccc>{\columncolor{mygray}}c}
\toprule
\textbf{Model} & ARES & AREC & ATTR & CNT & ISYN & OCR & ORES & OREC & SPER & SRES & TPER & TRES & Avg. \\
\midrule
\multicolumn{14}{l}{\textbf{\textit{Base Models}}} \\
Qwen3-VL-8B~\citep{qwen3vl} & 62.2 & 54.0 & 51.9 & 43.8 & 68.1 & 42.9 & 62.9 & 57.4 & 33.3 & 45.5 & 33.3 & 67.0 & 61.0 \\
Gemini 2.5 Pro~\citep{gemini2.5} & 56.9 & 47.6 & 66.7 & 41.7 & 71.8 & \textbf{57.1} & 53.3 & 40.7 & 0.0 & \textbf{72.7} & \textbf{66.7} & 48.4 & 55.7 \\
GPT-5~\citep{gpt5} & 71.1 & 69.8 & \textbf{70.4} & 47.9 & \textbf{88.3} & \textbf{57.1} & \textbf{75.8} & 74.1 & 33.3 & \textbf{72.7} & 50.0 & 75.8 & 74.3 \\
\midrule
\multicolumn{14}{l}{\textbf{\textit{Long Video LLMs}}} \\
VideoChat-Flash~\citep{videochat-flash} & 35.0 & 42.9 & 37.0 & 31.3 & 34.4 & 42.9 & 60.0 & 46.3 & 33.3 & 54.5 & 33.3 & 46.2 & 44.1 \\
Time-R1~\citep{time-r1} & 20.6 & 28.6 & 25.9 & 35.4 & 31.9 & 35.7 & 53.3 & 48.2 & 33.3 & 36.4 & 50.0 & 44.0 & 37.6 \\
Video-RTS~\citep{video-rts} & 43.3 & 52.4 & 40.7 & 39.6 & 33.7 & 42.9 & 60.8 & 53.7 & 33.3 & 45.5 & 50.0 & 49.5 & 47.9 \\
\midrule
\multicolumn{14}{l}{\textbf{\textit{RAG-based Video LLMs}}} \\
LightRAG~\citep{lightrag} & 41.7 & 30.2 & 40.7 & 35.4 & 54.0 & 50.0 & 46.7 & 61.1 & 33.3 & 45.5 & 50.0 & 52.8 & 46.6 \\
HippoRAG~\citep{hipporag} & 45.6 & 47.6 & 40.7 & 37.5 & 52.2 & 42.9 & 52.9 & 64.8 & \textbf{66.7} & 54.5 & 50.0 & 70.3 & 52.1 \\
Video-RAG~\citep{video-rag} & 51.7 & 47.6 & 37.0 & 39.6 & 49.7 & \textbf{57.1} & 62.1 & 68.5 & \textbf{66.7} & 45.5 & 50.0 & 68.1 & 55.4 \\
\midrule
\multicolumn{14}{l}{\textbf{\textit{Memory-based Video LLMs}}} \\
EgoRAG~\citep{egolife} & 31.1 & 55.6 & 33.3 & 22.9 & 41.1 & 28.6 & 44.6 & 48.2 & 33.3 & 54.5 & \textbf{66.7} & 48.4 & 41.1 \\
Ego-R1~\citep{ego-r1} & 37.2 & 52.4 & 40.7 & 35.4 & 38.0 & 35.7 & 42.1 & 51.9 & \textbf{66.7} & 63.6 & 50.0 & 52.8 & 42.7 \\
HippoMM~\citep{hippomm} & 41.1 & 42.9 & 55.6 & 35.4 & 38.7 & 35.7 & 37.9 & 53.7 & 33.3 & 54.5 & 50.0 & 47.3 & 41.6 \\
M3-Agent~\citep{m3-agent} & 52.2 & 57.1 & 59.3 & 45.8 & 51.5 & 42.9 & 54.6 & 64.8 & 33.3 & 45.5 & 50.0 & 71.4 & 55.3 \\
\midrule
\rowcolor{myblue} \multicolumn{14}{l}{\textbf{\textit{\modelname{} (Ours)}}} \\
\rowcolor{myblue} \textbf{\modelname{}-8B} & 65.0 & 66.7 & 59.3 & 41.7 & 72.4 & 42.9 & 67.5 & 72.2 & 33.3 & 54.5 & \textbf{66.7} & 69.2 & 66.0 \\
\rowcolor{myblue} \textbf{\modelname{}-GPT} & \textbf{81.1} & \textbf{73.0} & \textbf{70.4} & \textbf{54.2} & 85.3 & 42.9 & 75.0 & \textbf{77.8} & 33.3 & \textbf{72.7} & \textbf{66.7} & \textbf{79.1 }& \textbf{76.6 }\\
\bottomrule
\end{tabular}
}
\end{table*}

%% file: fig/fig_appendix_memtype_hippovlog.tex
\begin{figure*}[t!]
    \centering
    \includegraphics[width=0.6\linewidth]{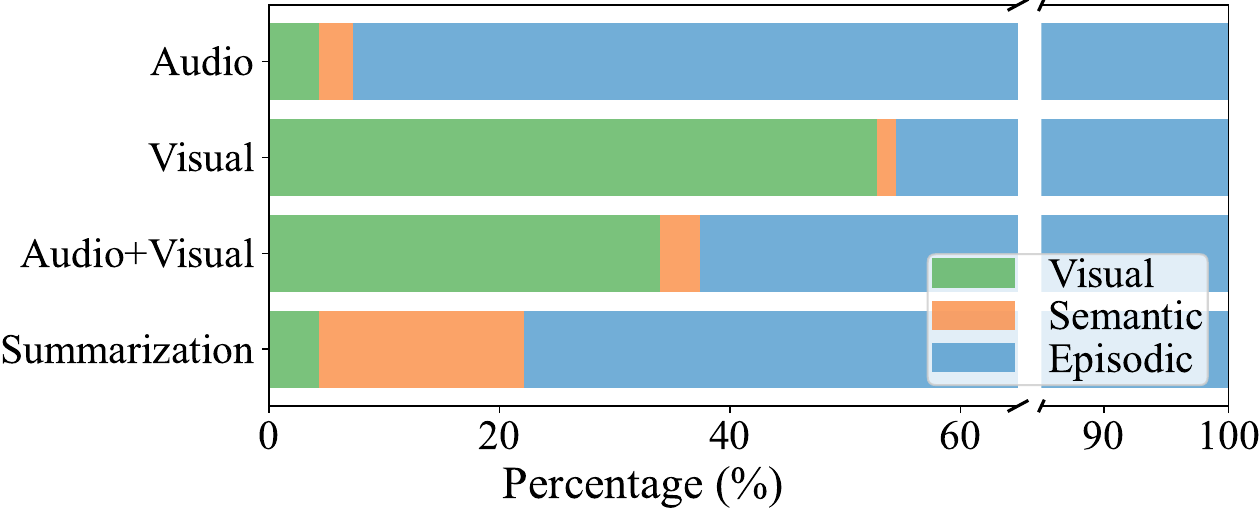}
    \caption{Memory type utilization of \modelname{} on four distinctive categories in HippoVlog.} 
    \label{fig:memtype_hippovlog}
\end{figure*}

%% file: tab/tab_dynamic_memory.tex
\begin{table*}[t]
\centering
\setlength{\tabcolsep}{3pt}
\renewcommand{\arraystretch}{1.1}
\caption{Category-wise average tIoU (\%) breakdown of \modelname{} and dynamic temporal scope retrieval baselines.}
\label{tab:appendix_tiou}
\resizebox{0.9\textwidth}{!}{
\begin{tabular}{lccccc>{\columncolor{mygray}}c ccccc>{\columncolor{mygray}}c ccc>{\columncolor{mygray}}c}
\toprule
\multirow{2}{*}{\textbf{Model}} &
\multicolumn{6}{c}{\textbf{EgoLifeQA}} &
\multicolumn{6}{c}{\textbf{Ego-R1 Bench}} &
\multicolumn{4}{c}{\textbf{LVBench}} \\
\cmidrule(lr){2-7} \cmidrule(lr){8-13} \cmidrule(lr){14-17}
 & Ent. & EvR. & Hab. & Rel. & Task & Avg.
 & Ent. & EvR. & Hab. & Rel. & Task & Avg.
 & Short & Med. & Long & Avg. \\
\midrule
Time-R1~\citep{time-r1} & 0.34 & 0.72 & 1.07 & 0.52 & 0.41 & 0.58 & 0.27 & 0.84 & 0.71 & 1.15 & 1.58 & 0.59 & 3.10 & 2.60 & 1.00 & 2.70 \\
\midrule
Qwen3 Emb.~\citep{qwen3embedding} & 2.87 & 4.31 & 5.58 & 2.98 & 8.91 & 4.35 & 2.68 & 2.74 & 3.85 & 2.74 & 3.70 & 2.87 & 4.48 & 6.20 & 1.75 & 4.54 \\
HippoRAG~\citep{hipporag} & 3.02 & 4.19 & 4.99 & 2.12 & 8.36 & 4.00 & 3.32 & 2.85 & 3.28 & 2.23 & 4.07 & 3.28 & 4.23 & 5.76 & 1.88 & 4.30 \\
InternVideo2~\citep{internvideo2} & 2.09 & 4.42 & 6.04 & 2.00 & 3.88 & 3.36 & 2.71 & 2.55 & 3.09 & 1.85 & 2.32 & 2.60 & 3.66 & 4.71 & 0.87 & 3.55 \\
\midrule
EgoRAG~\citep{egolife} & 3.20 & 3.38 & 4.62 & 3.10 & 4.82 & 3.60 & 2.40 & 3.07 & 4.08 & 2.19 & 3.78 & 2.73 & 4.10 & 3.38 & 0.91 & 3.50 \\
Ego-R1~\citep{ego-r1} & 3.31 & 3.52 & 5.03 & 2.87 & 5.18 & 3.70 & 2.57 & 2.83 & 4.13 & 2.83 & 4.12 & 2.89 & 4.08 & 3.72 & 1.14 & 3.60 \\
\midrule
AKS~\citep{aks} & 2.42 & 2.77 & 3.08 & 2.93 & 2.67 & 2.75 & 2.03 & 2.48 & 2.99 & 2.58 & 3.04 & 2.30 & 3.81 & 4.11 & 1.10 & 3.52 \\
\midrule
\rowcolor{myblue} \textbf{\modelname{} (Ours)} & \textbf{9.79} & \textbf{10.43} & \textbf{11.85} & \textbf{7.73} & \textbf{12.97} & \textbf{10.09} & \textbf{8.91} & \textbf{9.85} & \textbf{8.86} & \textbf{9.63} & \textbf{9.58} & \textbf{9.17} & \textbf{7.53} & \textbf{14.41} & \textbf{10.02} & \textbf{9.57} \\
\bottomrule
\end{tabular}
}
\end{table*}

\begin{table*}[t]
\centering
\setlength{\tabcolsep}{3pt}
\renewcommand{\arraystretch}{1.1}
\caption{Category-wise performance breakdown of \modelname{} and dynamic temporal scope retrieval baselines.}
\label{tab:appendix_dyna_perf}
\resizebox{0.9\textwidth}{!}{
\begin{tabular}{lccccc>{\columncolor{mygray}}c ccccc>{\columncolor{mygray}}c ccc>{\columncolor{mygray}}c}
\toprule
\multirow{2}{*}{\textbf{Model}} &
\multicolumn{6}{c}{\textbf{EgoLifeQA}} &
\multicolumn{6}{c}{\textbf{Ego-R1 Bench}} &
\multicolumn{4}{c}{\textbf{LVBench}} \\
\cmidrule(lr){2-7} \cmidrule(lr){8-13} \cmidrule(lr){14-17}
 & Ent. & EvR. & Hab. & Rel. & Task & Avg.
 & Ent. & EvR. & Hab. & Rel. & Task & Avg.
 & Short & Med. & Long & Avg. \\
\midrule
Time-R1~\citep{time-r1} & 39.2 & 50.8 & 65.6 & 48.8 & 47.6 & 48.8 & 49.2 & 48.8 & 46.2 & 42.1 & 44.7 & 48.0 & 32.1 & 23.6 & 40.2 & 31.1 \\
\midrule
Qwen3 Emb.~\citep{qwen3embedding} & 44.0 & 59.5 & 70.5 & 58.4 & 68.3 & 57.8 & 51.9 & 65.9 & 61.5 & 57.9 & 47.4 & 54.0 & 52.9 & 49.1 & 62.3 & 53.2 \\
HippoRAG~\citep{hipporag} & 48.8 & 60.3 & 70.5 & 60.8 & 66.7 & 59.6 & 54.5 & 65.9 & 69.2 & 52.6 & 50.0 & 56.0 & 54.9 & 47.5 & 62.3 & 54.0 \\
InternVideo2~\citep{internvideo2} & 40.8 & 54.0 & 60.7 & 51.2 & 52.4 & 50.6 & 50.3 & 56.1 & 46.2 & 47.4 & 52.6 & 51.0 & 47.4 & 37.3 & 53.4 & 45.7 \\
\midrule
EgoRAG~\citep{egolife} & 40.0 & 56.3 & 62.3 & 54.4 & 52.4 & 52.0 & 46.6 & 56.1 & 46.2 & 47.4 & 55.3 & 49.0 & 32.4 & 32.0 & 31.9 & 32.2 \\
Ego-R1~\citep{ego-r1} & 51.2 & 53.2 & 63.9 & 50.4 & 50.8 & 53.0 & 50.8 & 63.4 & 38.5 & 36.8 & 57.9 & 52.0 & 32.5 & 36.5 & 37.3 & 34.1 \\
\midrule
AKS~\citep{aks} & 41.6 & 51.6 & 63.9 & 51.2 & 52.4 & 50.6 & 51.3 & 63.4 & 46.2 & 36.8 & 50.0 & 51.7 & 43.3 & 33.9 & 39.2 & 40.4 \\
\midrule
\rowcolor{myblue} \textbf{\modelname{} (Ours)} & \textbf{62.4} & \textbf{64.3} & \textbf{75.4} & \textbf{62.4} & \textbf{71.4} & \textbf{65.6} & \textbf{64.6} & \textbf{70.7} & \textbf{76.9} & \textbf{57.9} & \textbf{63.2} & \textbf{65.3} & \textbf{58.3} & \textbf{65.4} & \textbf{72.1} & \textbf{61.9} \\
\bottomrule
\end{tabular}
}
\end{table*}

%% file: tab/tab_backbones.tex
\begin{table}[h]
\centering
\setlength{\tabcolsep}{2pt}
\caption{Performance of \modelname{} with various backbones.}
\label{tab:diff_backbones}
\resizebox{\linewidth}{!}{
\begin{tabular}{lccc}
\toprule
\textbf{Model} & \textbf{EgoLifeQA} & \textbf{LVBench} & \textbf{VideoMME (L)} \\
\midrule
WorldMM-Gemini + Qwen3-VL-Emb & 67.4 & 61.5 & 74.9 \\
WorldMM-Gemini + VLM2Vec-V2 & \textbf{68.2} & 61.7 & 75.8 \\
WorldMM-GPT + Qwen3-VL-Emb & 66.0 & 61.4 & 75.8 \\
\rowcolor{myblue} WorldMM-GPT + VLM2Vec-V2 & 65.6 & \textbf{61.9} & \textbf{76.6} \\
\bottomrule
\end{tabular}
}
\end{table}

%% file: tab/tab_temp_scale.tex
\begin{table}[h]
\centering
\setlength{\tabcolsep}{12pt}
\caption{Performance with different episodic timescales.}
\label{tab:temp_scale}
\resizebox{0.5\linewidth}{!}{
\begin{tabular}{lc}
\toprule
\textbf{Temporal Scale} & \textbf{Acc} \\
\midrule
20s/2m/5m/50m & 65.2 \\
\rowcolor{myblue} 30s/3m/10m/1h & 65.6 \\
1m/5m/15m/1.5h & 64.8 \\
\bottomrule
\end{tabular}
}
\end{table}

%% file: tab/qualitative/tab_episodic_triple_extraction.tex
\begin{table*}[t]
    \centering
    \renewcommand{\arraystretch}{1.0}
    \renewcommand{\tabcolsep}{3mm}
    \caption{Example of episodic triplet extraction.}
    \small
    \begin{tabular}{l p{0.7\linewidth}}
        \toprule
        \textbf{Caption} & I stand and walk to the other side of the dining table. Katrina asks, “Is this for tomorrow’s game?” “Yes—let’s think about what to do tomorrow,” I say. I raise my right hand as Katrina walks toward me. Lucia asks, “Using ancient poems? Or what else?” Katrina says, “I’m not good with ancient poems.” Tasha asks, “Then what else to use?” Katrina says, “I’ll be out in the first round. My room is already cleaned up.” “Okay,” I say. I turn toward the stairs, put down my phone, look back at the living room door, and walk into the second-floor living room. Lucia adds, “For example, not coming out.” Katrina says, “Let me check that place we’re going to.” Tasha asks, “I just want to ask which fields it has expanded into.” Lucia says, “Okay.” \\
        \midrule
        \textbf{Extracted Triplets}
& [I, stand at, dining table]\\
& [I, walk to, other side of the dining table]\\
& [Katrina, asks about, tomorrow]\\
& [I, confirm, tomorrow]\\
& [I, raise, right hand]\\
& [Katrina, walks toward, I]\\
& [Lucia, asks about, using ancient poems]\\
& [Katrina, says, not good with ancient poems]\\
& [Tasha, asks, what else to use]\\
& [Katrina, says, I will be out in the first round]\\
& [Katrina, has, room already cleaned up]\\
& [I, turn toward, stairs]\\
& [I, put down, phone]\\
& [I, look back at, living room door]\\
& [I, walk into, second-floor living room]\\
& [Lucia, adds, not coming out as an example]\\
& [Katrina, says, let me check that place we're going to]\\
& [Lucia, says, Okay]\\
        \bottomrule
    \end{tabular}
    \label{tab:episodic_triple_extraction}
\end{table*}

%% file: tab/qualitative/tab_semantic_triple_extraction.tex
\begin{table*}[t]
    \centering
    \renewcommand{\arraystretch}{1.0}
    \renewcommand{\tabcolsep}{3mm}
    \caption{Example of semantic triplet extraction.}
    \small
    \begin{tabular}{l p{0.7\linewidth}}
        \toprule
        \textbf{Caption} & I got up, moved my phone, and checked it before turning it off. Alice expressed her feelings towards me, and I responded by checking my phone's chat interface. Alice then questioned her appearance, and I turned off the phone, looking around at the snacks and utensils on the table. I stood up, grabbed a pack of snacks, and proceeded to my room to enjoy them. Alice asked about something being fancy, and I fetched my glasses, placing them on the table. ... I managed my phone, swiping through pages, and interacted with others as I went about my tasks. I observed Alice and Tasha, discussing what to feed a cat, and continued interacting with my phone. As the environment darkened, I engaged with the surroundings, noting the layout and structures. Finally, I moved towards a house with blue-green walls, managing my power bank and surveying the area. \\
        \midrule
        \textbf{Extracted Triplets}
& [I, assigns tasks to, Katrina]\\
& [I, handles reimbursements for, Alice]\\
& [I, uses WeChat for, money transfers]\\
& [I, often eats, snacks]\\
& [I, wears, glasses]\\
& [Lucia, dislikes, overly sweet food]\\
& [Alice, expresses romantic feelings toward, I]\\
& [Katrina, helps with, expense tracking]\\
& [I, requires PDFs for, reimbursement]\\
& [Tasha, participates in, house demolition tasks]\\
& [Lucia, participates in, house demolition tasks]\\
        \bottomrule
    \end{tabular}
    \label{tab:semantic_triple_extraction}
\end{table*}

%% file: tab/qualitative/tab_semantic_triple_consolidation.tex
\begin{table*}[t]
    \centering
    \renewcommand{\arraystretch}{1.0}
    \renewcommand{\tabcolsep}{3mm}
    \caption{Example of semantic consolidation.}
    \small
    \begin{tabular}{l l r}
        \toprule
        \textbf{Original Triplets}
& [I, uses WeChat to send money]\\
& [I, wears, glasses]\\
& [I, often eats, fruits]\\
& [Lucia, likes, sweet desserts]\\
& [Tasha, participates in, household projects]\\
        \midrule
        \textbf{New Triplets}
& [I, assigns tasks to, Katrina]\\
& [I, handles reimbursements for, Alice]\\
& [I, uses WeChat for, money transfers]\\
& [I, often eats, snacks]\\
& [I, wears, glasses]\\
& [Lucia, dislikes, overly sweet food] & \textcolor{gray}{\textit{\% conflicts with existing ``likes sweet desserts"}}\\
& [Alice, expresses romantic feelings toward, I]\\
& [Katrina, helps with, expense tracking]\\
& [I, requires PDFs for, reimbursement]\\
& [Tasha, participates in, house demolition tasks]\\
& [Lucia, participates in, house demolition tasks]\\
        \midrule
        \textbf{Consolidated Triplets}
& [I, assigns tasks to, Katrina]\\
& [I, handles reimbursements for, Alice]\\
& [I, uses,  WeChat to send money] & \textcolor{gray}{\textit{\% merged with ``uses WeChat for money transfers"}}\\
& [I, often eats, fruits and snacks] & \textcolor{gray}{\textit{\% merged and generalized ``fruits" and ``snacks"}}\\
& [I, wears, glasses] & \textcolor{gray}{\textit{\% merged with duplicate in new triples}}\\
& [Alice, expresses romantic feelings toward, I]\\
& [Katrina, helps with, expense tracking]\\
& [I, requires PDFs for, reimbursement]\\
& [Tasha, participates in, house demolition tasks] & \textcolor{gray}{\textit{\% refined general triple ``household projects"}}\\
& [Lucia, participates in, house demolition tasks]\\
        \bottomrule
    \end{tabular}
    \label{tab:semantic_triple_consolidation}
\end{table*}

%% file: tab/qualitative/tab_multi_turn_refinement.tex
\begin{table*}[t]
    \centering
    \renewcommand{\arraystretch}{1.0}
    \renewcommand{\tabcolsep}{3mm}
    \caption{Example of the multi-turn refinement of \modelname{}.}
    \small
    \begin{tabular}{l p{0.7\linewidth}}
        \toprule
        \textbf{Question} & What were we doing last time we discussed the air conditioning temperature? \\
        & (A) Eating hot pot (B) Shopping (C) Playing piano (D) Picking up a package \\
        \midrule
        \textbf{Answer} & A \\
        \midrule
        \textbf{Round 1} & Decision: Search // Memory: Episodic \\
        & Search Query: \textcolor{blue}{discussing the air conditioning temperature} \\
        & \textit{Retrieved:} \\
        & \textit{[DAY2 13:36:00-13:39:00]} \\
        & \textit{I joined my colleagues at the dining table and engaged in a discussion about the food while preparing to eat. ... During the meal, we discussed the air conditioner situation in the rooms. Tasha and Katrina talked about where the remotes were and who had access, with Shure mentioning that his room's AC had been on continuously due to a borrowed remote from me. I listened to their conversation, ...} \\
        & \textit{[DAY2 17:48:00-17:51:00]} \\
        & \textit{I coordinated with Shure and Lucia to find a suitable spot for my power bank, initially placing it on a flower pot as Shure suggested, but realizing it was unstable. ... We discussed the room environment, including turning off the air conditioner that was blowing cold air. Shure clarified that the remote we had was for the TV, not the air conditioner, prompting me to inquire about the correct remote. ...} \\
        \midrule
        \textbf{Round 2} & Decision: Search // Memory: Episodic \\
        & Search Query: \textcolor{blue}{air conditioning} \\
        & \textit{Retrieved:} \\
        & \textit{[DAY2 17:48:30-17:49:00]} \\
        & \textit{``Flip it and put it underneath," I say. ... ``Wow, this, this…" Lucia murmurs. ``How do you turn off this air conditioner? It’s blowing so cold up here." I adjust the power bank for a better position. Shure says, ``Just tell it to turn off, it’ll turn off." ``Really? It’s voice-controlled?" Lucia asks. I secure the power bank and turn to Shure and Lucia. Shure adds, ``No, this is the TV remote." ``Where is it?" I ask, and I chat with them.} \\
        & \textit{[DAY2 18:34:01-18:34:29]} \\
        & \textit{I watch Shure set the air conditioner to 26 degrees. ``Save the Earth," I say. ``Everyone’s responsibility," I laugh. I kneel, then sit on the sofa. Shure asks, ``Seriously, you think it’s good to record this?" I look at my friends chatting and watch the movie. Shure says, ``We’re still using the AC." Shure says, ``He really can’t hear a thing." I watch Shure take a few steps. I laugh. Shure says, ``Hahaha, it’s like we’re watching a silent movie." I smile and keep watching the movie. ``Whoa, this is classic," I say. I stand up. ``Come on, let’s have a drink," I say.} \\
        & \textit{[DAY2 18:33:00-18:33:30]} \\
        & \textit{I walk in front of the fill light, pass behind Choiszt, and move toward the curtain. ... ``The popularity is always high," I say. Shure: ``Ha." Choiszt: ``Turn it on." ``Turn on the air conditioner?" I ask. Lucia: ``I'll just, I suggest I drink water directly from a bowl." Lucia: ``So later I'll eat with this bowl."} \\
        \midrule
        \textbf{Round 3} & Decision: Search // Memory: \textcolor{blue}{Visual} \\
        & Search Query: DAY2 18:34:01-18:34:29 \\
        & \textit{Retrieved:} \vspace{0.2em} \\
        & \filmbox{'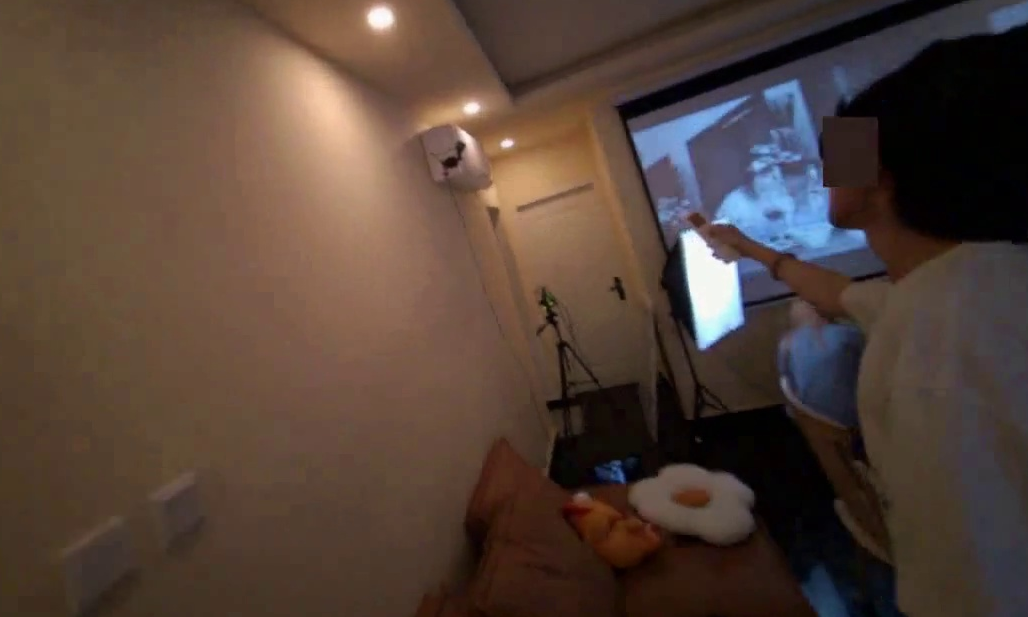'}\filmbox{'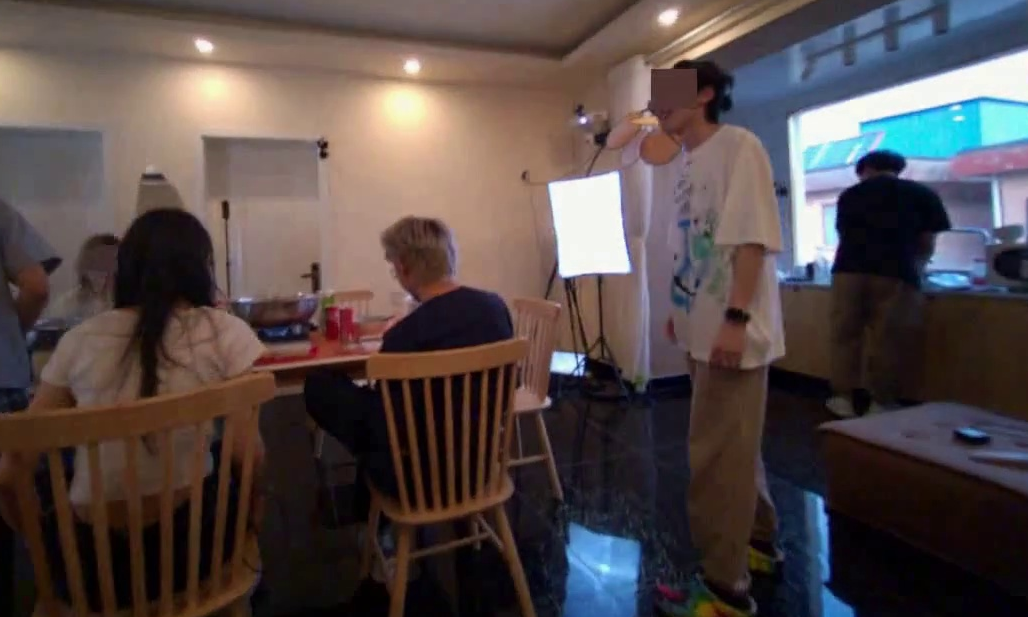'}\filmbox{'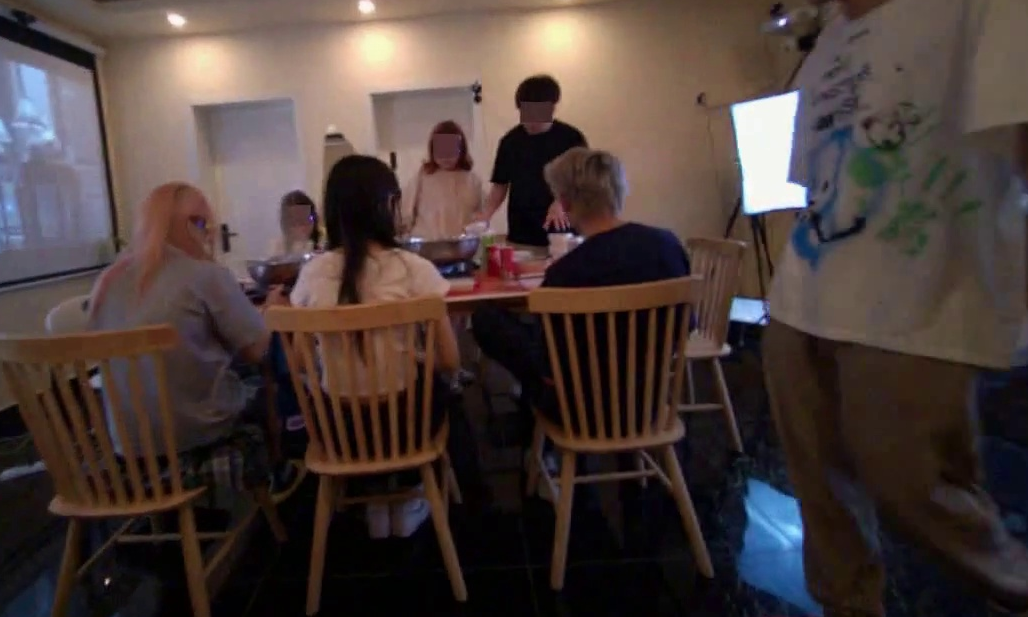'}\filmbox{'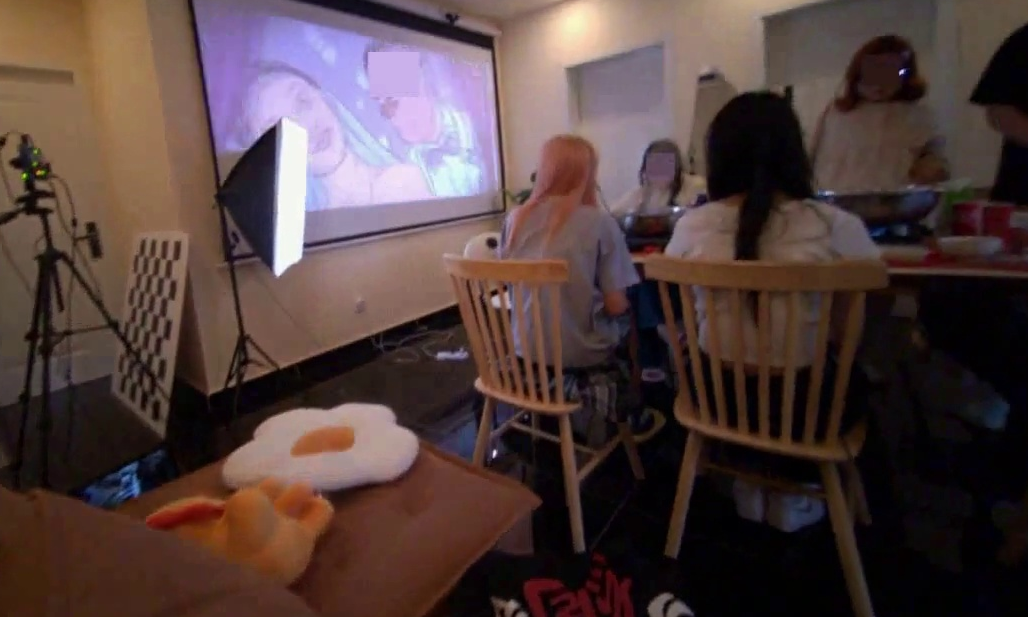'}\filmbox{'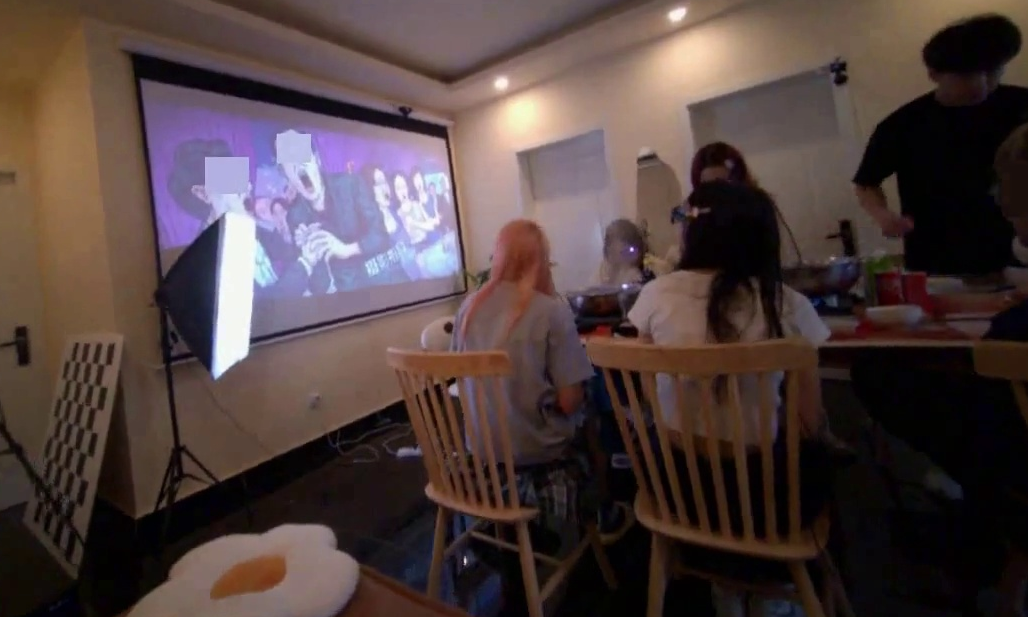'}\vspace{0.2em} \\
        & \textit{(Author Note: They are gathered around a table cooking and sharing a hot pot meal while watching a projection on the wall.)} \\
        \midrule
        \textbf{Round 4} & Decision: Answer \\
        \midrule
        \textbf{Response} & A \\
        \bottomrule
    \end{tabular}
    \label{tab:qualitative_multi_turn_refinement}
\end{table*}

%% file: fig/prompts/fig_prompt_video_caption.tex
\begin{figure*}[t]
\centering
\begin{tcolorbox}[colback=white, colframe=black, boxrule=0.5mm, arc=2mm, left=2mm, right=2mm, top=2mm, bottom=2mm]
You are an expert video captioner.\\

You will receive a short video segment represented by ordered frames and optional transcript lines with timestamps.
Write a caption describing both the visual content and the audible content of the segment.\\

Guidelines:\\
- Describe visible actions, people, objects, and environment.\\
- Include relevant speech, sounds, or audio events.\\
- Keep the caption factual and neutral.\\
- Do not mention frames, timestamps, or that the input came from frames.\\
- Avoid speculation about emotions or intentions unless clearly visible or stated in speech.\\

Output only the final caption text.
\end{tcolorbox}
\caption{Prompt for video caption generation.}
\label{fig:prompt_video_caption}
\end{figure*}

%% file: fig/prompts/fig_prompt_episodic_ner_extraction.tex
\begin{figure*}[t]
\centering
\begin{tcolorbox}[colback=white, colframe=black, boxrule=0.5mm, arc=2mm, left=2mm, right=2mm, top=2mm, bottom=2mm]

Your task is to extract named entities from the given paragraph. Respond with a JSON list of entities.\\

Example:\\

Radio City is India's first private FM radio station and was started on 3 July 2001.
It plays Hindi, English and regional songs.
Radio City recently forayed into New Media in May 2008 with the launch of a music portal - PlanetRadiocity.com that offers music related news, videos, songs, and other music-related features.\\

\{
``named\_entities":\\
\hspace*{2em}[``Radio City'', ``India'', ``3 July 2001'', ``Hindi'', ``English'', ``May 2008'', ``PlanetRadiocity.com'']\\
\}
\end{tcolorbox}
\vspace{-0.1in}
\caption{Prompt for named entity recognition (NER). Recognized named entities are used to extract episodic triplets as shown in \cref{fig:prompt_episodic_triple_extraction}.}
\label{fig:prompt_episodic_ner_extraction}
\end{figure*}

%% file: fig/prompts/fig_prompt_episodic_triple_extraction.tex
\begin{figure*}[t]
\centering
\begin{tcolorbox}[colback=white, colframe=black, boxrule=0.5mm, arc=2mm, left=2mm, right=2mm, top=2mm, bottom=2mm]
Your task is to construct an RDF (Resource Description Framework) graph from the given passages and named entity lists. 
Respond with a JSON list of triples, with each triple representing a relationship in the RDF graph.\\

Pay attention to the following requirements:
\begin{itemize}
    \item Each triple should contain at least one, but preferably two, of the named entities in the list for each passage.
    \item When resolving pronouns, if the pronoun refers to the first-person (e.g., I, me, my), keep it as ``I" instead of replacing with terms like ``speaker" or ``narrator". For other pronouns, clearly resolve them to their specific names to maintain clarity.
\end{itemize}
\vspace{1em}
Convert the paragraph into a JSON dict, it has a named entity list and a triple list.\\

Example:\\

Radio City is India's first private FM radio station and was started on 3 July 2001.
It plays Hindi, English and regional songs.
Radio City recently forayed into New Media in May 2008 with the launch of a music portal - PlanetRadiocity.com that offers music related news, videos, songs, and other music-related features.\\

\{
``named\_entities":\\
\hspace*{2em}[``Radio City'', ``India'', ``3 July 2001'', ``Hindi'', ``English'', ``May 2008'', ``PlanetRadiocity.com'']\\
\}

\vspace{1em}
\{
``triples": [\\
\hspace*{2em}[``Radio City", ``located in", ``India"],\\
\hspace*{2em}[``Radio City", ``is", ``private FM radio station"],\\
\hspace*{2em}[``Radio City", ``started on", ``3 July 2001"],\\
\hspace*{2em}[``Radio City", ``plays songs in", ``Hindi"],\\
\hspace*{2em}[``Radio City", ``plays songs in", ``English"],\\
\hspace*{2em}[``Radio City", ``forayed into", ``New Media"],\\
\hspace*{2em}[``Radio City", ``launched", ``PlanetRadiocity.com"],\\
\hspace*{2em}[``PlanetRadiocity.com", ``launched in", ``May 2008"],\\
\hspace*{2em}[``PlanetRadiocity.com", ``is", ``music portal"],\\
\hspace*{2em}[``PlanetRadiocity.com", ``offers", ``news"],\\
\hspace*{2em}[``PlanetRadiocity.com", ``offers", ``videos"],\\
\hspace*{2em}[``PlanetRadiocity.com", ``offers", ``songs"]\\
]\}
\end{tcolorbox}
\vspace{-0.1in}
\caption{Prompt for episodic triplet extraction.}
\label{fig:prompt_episodic_triple_extraction}
\end{figure*}

%% file: fig/prompts/fig_prompt_episodic_multi_construction.tex
\begin{figure*}[t]
\centering
\begin{tcolorbox}[colback=white, colframe=black, boxrule=0.5mm, arc=2mm, left=2mm, right=2mm, top=2mm, bottom=2mm]
As an Event Summary Documentation Specialist, your role is to systematically structure and summarize event information, ensuring that all key actions of major characters are captured while maintaining clear event logic and completeness. Your focus is on concise and factual summarization rather than detailed transcription.\\

\# Specific Requirements\\

1. Structure the Events Clearly\\
- Merge related events: Consolidate similar content into major events and arrange them in chronological order to ensure a smooth logical flow.\\
- Logical segmentation: Events can be grouped based on location, task, or theme. Each event should have a clear starting point, progression, and key turning points without any jumps or fragmentation in the information.\\
\\
2. Retain Key Information\\
- All subjects' decisions and actions must be fully presented, including all critical first-person activities. Transitions between different parts, such as moving between floors or starting/ending a task, should be seamless.\\
- Any discussions, decisions, and task execution involving the primary character and other key individuals that impact the main storyline must be reflected. This includes recording, planning, and confirming matters, but in a concise manner.\\
- The purpose and method of key actions must be recorded, such as ``ordering takeout using a phone" or ``documenting a plan on a whiteboard."\\
\\
3. Concise Expression, Remove Redundancies\\
- Keep the facts clear, avoiding descriptions of atmosphere, emotions, or abstract content.\\
- Remove trivial conversations and extract only the core topics and conclusions of discussions. If a discussion is lengthy, summarize it into task arrangements, decision points, and specific execution details.\\
\\
4. Strictly Adhere to Facts, No Assumptions\\
- Do not make assumptions or add interpretations—strictly organize content based on available information, ensuring accuracy. Every summarized point must have a basis in the original information, with no unnecessary additions.\\
- Maintain the correct chronological order of events. The sequence of developments must strictly follow their actual occurrence without any inconsistencies.\\
\\
\# Output Format\\
Each paragraph should represent one major event, structured in a summary-detail-summary format. Strictly output below \{word\_limit\} words in total. Do not report the word count in the output.
\end{tcolorbox}
\caption{Prompt for episodic memory construction to generate coarser-level caption.}
\label{fig:prompt_episodic_multi_construction}
\end{figure*}

%% file: fig/prompts/fig_prompt_episodic_multi_retrieval.tex
\begin{figure*}[t]
\centering
\begin{tcolorbox}[colback=white, colframe=black, boxrule=0.5mm, arc=2mm, left=2mm, right=2mm, top=2mm, bottom=2mm]
You are an expert assistant that helps filter and select relevant video captions based on a given query.
Your task is to analyze the retrieved video captions and determine which ones are most relevant to answer the question.\\

Given the following question and retrieved video captions, select and rank the most relevant captions that should be used to answer the question.\\

Instructions:\\
1. Consider the nature of the question when selecting captions:\\
\hspace*{1em}- e.g., for queries about specific events, focus on finer granularities; for habitual, relationship, or general queries, consider coarser granularities.\\
\hspace*{1em}- Note that coarser granularity captions may provide broader context, but finer granularity captions often contain more specific details.\\
2. Each caption shows its time range (start\_time to end\_time)\\
3. Analyze each caption for relevance to the question\\
4. Select captions that directly help answer the question\\
5. Return the IDs in ranked order (most relevant first)\\
6. Only include captions that are truly relevant\\

Return ONLY a JSON array of caption IDs in order of relevance (most relevant first), without additional justification.
\end{tcolorbox}
\caption{Prompt for episodic memory retrieval to select from multiple timescales.}
\label{fig:prompt_episodic_multi_retrieval}
\end{figure*}

%% file: fig/prompts/fig_prompt_semantic_triple_extraction.tex
\begin{figure*}[t]
\centering
\begin{tcolorbox}[colback=white, colframe=black, boxrule=0.5mm, arc=2mm, left=2mm, right=2mm, top=2mm, bottom=2mm]
\footnotesize
You are tasked with extracting semantic knowledge from episodic triples. 
Your goal is to infer generalizable information that extends beyond the specific episode. 
Focus on capturing valid semantic triples that can guide reasoning about behavior, relationships, or preferences.\\

\# What to Extract\\
1. Relationships: social bonds or roles between entities  that persist over time\\
\hspace*{1em}(e.g., ``Alice is a friend with Bob", ``Jason is a teacher of Alice").\\
2. Attributes \& Preferences: tendencies, likes/dislikes, personality-like traits, or behavioral habits\\
\hspace*{1em}(e.g., ``Alice prefers not having dessert", ``Bob enjoys music").\\
3. Habits \& Capabilities: actions or patterns that suggest what an entity often does, can do, or tends to do\\
\hspace*{1em}(e.g., ``Alice often helps friends", ``Jason can give advice").\\
4. Conceptual Knowledge: directly useful facts that support reasoning, but avoid overly broad taxonomic statements\\
\hspace*{1em}(e.g., ``Alice's office is near Cafe X", ``Bob's gym is closed on Sundays").\\

\# What to Avoid\\
- One-off events or transient states (e.g., “ate pizza yesterday”, “was late once”) unless explicitly declared as a preference/role\\
- Broad taxonomy or trivia unrelated to behavior (e.g., “a laptop is electronics”, “Paris is in France”)\\
- Speculative or mind-reading inferences without textual support (e.g., motives, beliefs not evidenced)\\

\# Important Notes\\
- Prefer to base semantic triples on multiple supporting episodes.\\
- BUT if a single episode clearly reflects a role, preference, habit, or capability, it is valid to include it.\\
- Each semantic triple MUST have at least one supporting episodic triple.\\
- Reduce duplication. If multiple episodic triples support the same or very similar semantic knowledge, merge them into one semantic triple rather than repeating.\\
- The `episodic\_evidence[i]' list must always point to the indices that support `semantic\_triples[i]'.\\
- Aim for broad coverage: extract as many valid semantic triples as reasonably supported by the input.\\

\# Output Format\\
- Return ONLY a JSON object with the following two keys:\\
\hspace*{1em}- `semantic\_triples' (List[List[str]]): Each item is a triple [subject, predicate, object].\\
\hspace*{1em}- `episodic\_evidence' (List[List[int]]) : Each item is a list of 0-based indices pointing to the input episodic triples that support the corresponding semantic triple at the same position.\\
- The two lists MUST have the same length and aligned order.\\
- If no semantic knowledge is inferable, return: \{``semantic\_triples": [], ``episodic\_evidence": []\}\\

Example:\\

Episodic triples: \\
0. [``Alice", ``talks to", ``Bob"], \\
1. [``Alice", ``laughs with", ``Bob"], \\
2. [``Alice", ``doesn't eat cake", ``at restaurant"], \\
3. [``Alice", ``shares personal stories with", ``Bob"], \\
4. [``Alice", ``brings coffee to", ``Bob"], \\
5. [``Jason", ``talks to", ``Alice"], \\
6. [``Alice", ``declines dessert", ``at friend's house"] \\

Output: \\
\{ \\
\hspace*{2em}``semantic\_triples": [ \\
\hspace*{4em}[``Alice", ``is a friend with", ``Bob"], \\
\hspace*{4em}[``Alice", ``prefers", ``not having dessert"] \\
\hspace*{2em}], \\
\hspace*{2em}``episodic\_evidence": [ \\
\hspace*{4em}[0, 1, 3], \\
\hspace*{4em}[2, 6] \\
\hspace*{2em}] \\
\}
\end{tcolorbox}
\vspace{-0.1in}
\caption{Prompt for semantic triplet extraction.}
\label{fig:prompt_semantic_triple_extraction}
\end{figure*}

%% file: fig/prompts/fig_prompt_semantic_consolidation.tex
\begin{figure*}[t]
\centering
\begin{tcolorbox}[colback=white, colframe=black, boxrule=0.5mm, arc=2mm, left=2mm, right=2mm, top=2mm, bottom=2mm]
You are tasked with consolidating semantic knowledge by processing a new semantic triple against relevant existing knowledge from previous timestamps.\\

Your job is to make two decisions:\\
1. Which existing triples to remove/pop — those that should be merged with the new triple or conflict with it\\
2. How to update the new triple — to capture merged information or resolve conflicts\\

\# Consolidation Rules\\
1. Merge Similar Information: If existing triples express very similar information to the new triple, remove them and update the new triple to contain the most complete/accurate form.\\
2. Resolve Conflicts: If the new triple conflicts with existing ones, decide which is more accurate/recent and remove outdated ones.\\
3. Update with Context: Use information from existing triples to make the new triple more specific or more accurate.\\
4. Preserve Unique Information: Only remove existing triples when they are redundant or conflicting.\\

\# Output Format\\
Return ONLY a JSON object with the following two keys:\\
- `updated\_triple' (List[str]): The new triple, possibly updated [subject, predicate, object].\\
- `triples\_to\_remove' (List[int]): Indices of existing triples to remove (empty list if none).\\

Example:\\

New triple: [``Alice", ``enjoys", ``coffee"]\\

Existing triples:\\
0. [``Alice", ``likes", ``beverages"]\\
1. [``Alice", ``favors", ``to have coffee after dinner"]\\
2. [``Alice", ``prefers", ``hot drinks"]\\
3. [``Alice", ``likes to drink", ``coffee"]\\[6pt]

Output:\\
\{\\
\hspace*{2em}``updated\_triple": [``Alice", ``likes", ``coffee"],\\
\hspace*{2em}``triples\_to\_remove": [1, 3]\\
\}
\end{tcolorbox}
\vspace{-0.1in}
\caption{Prompt for semantic memory consolidation.}
\label{fig:prompt_semantic_triple_consolidation}
\end{figure*}

%% file: fig/prompts/fig_prompt_retrieval_agent.tex
\begin{figure*}[t]
\centering
\begin{tcolorbox}[colback=white, colframe=black, boxrule=0.5mm, arc=2mm, left=2mm, right=2mm, top=2mm, bottom=2mm]
You are a reasoning agent for a video memory retrieval system.
Your job is to decide whether to stop and answer, or to search memory for more evidence.
When searching, you must select exactly one memory type and form a query.\\

\# Decision Modes\\
1. search: Retrieve memory to begin, continue, or extend progress toward the answer\\
   \hspace*{1em}- Choose one memory type and form a keyword(phrase)-style search query.\\
2. answer: Stop searching because the accumulated results are sufficient.\\
   \hspace*{1em}- No memory type selection is needed.\\

\# Memory Types\\
1. Episodic: Specific events/actions. Stores memories of past events and actions. Query by EVENT/ACTION.\\
2. Semantic: Entities/relationships. Stores factual knowledge about entities and their relationships, roles, and habits. Query by ENTITY/CONCEPT.\\
3. Visual: Scene/setting snapshots. Stores visual snapshots of scenes and settings. Query by SCENE/SETTING or TIMESTAMP RANGE.\\
    \hspace*{1em}- For timestamp range queries, return in the format: DAY X HH:MM:SS - DAY Y HH:MM:SS.

\# Context Inputs\\
- Current Query\\
- Round History: Log of past retrieval rounds. Each round is written in this format:\\

  \hspace*{2em}\#\#\# Round N\\
  \hspace*{2em}Decision: \textless search$|$answer\textgreater\\
  \hspace*{2em}Memory: \textless episodic$|$semantic$|$visual\textgreater\\
  \hspace*{2em}Search Query: \textless query text\textgreater\\
  \hspace*{2em}Retrieved: \textless retrieved items\textgreater\\

\# Strict Output Rules\\
- If decision = ``search": Must include ``selected\_memory" with exactly one memory type and one query.\\
- If decision = ``answer": Do NOT include ``selected\_memory".\\
- Always output in valid JSON only, no extra commentary.\\

\# Output Format\\
\{\\
 \hspace*{2em}``decision": ``search" $|$ ``answer",\\
 \hspace*{2em}``selected\_memory": \{\\
   \hspace*{4em}``memory\_type": ``episodic" $|$ ``semantic" $|$ ``visual",\\
   \hspace*{4em}``search\_query": \textless str\textgreater\\
 \hspace*{2em}\} \# Omit if decision = ``answer"\\
\}\\

(Few-shot examples given)
\end{tcolorbox}
\caption{Prompt for retrieval agent to decide retrieval strategy.}
\label{fig:prompt_retrieval_agent}
\end{figure*}

%% file: fig/prompts/fig_prompt_response_agent.tex
\begin{figure*}[t]
\centering
\begin{tcolorbox}[colback=white, colframe=black, boxrule=0.5mm, arc=2mm, left=2mm, right=2mm, top=2mm, bottom=2mm]
You are an AI assistant that answers questions about video using retrieved memory context. Your task is to answer multiple choice questions based on this accumulated context. Always choose the most relevant answer from the given choices based on the evidence provided.\\

\# Guidelines\\
- Analyze all provided context carefully.\\
- Choose the answer that best matches the evidence.\\
- If evidence is unclear, make the most reasonable inference.\\

\# Output Format\\
Provide your answer as a single letter (A, B, C, or D) based on the evidence.
\end{tcolorbox}
\caption{Prompt for response agent to generate response based on retrieved results.}
\label{fig:prompt_response_agent}
\end{figure*}